\documentclass{article}

\usepackage[preprint]{neurips_2022}

\usepackage{hyperref}       
\usepackage{url}            
\usepackage{booktabs}       
\usepackage{amsfonts,amsthm,amsmath}       
\usepackage{xcolor}         
\usepackage{bm}

\usepackage{algorithm}
\usepackage{algpseudocode}
\usepackage{pdflscape}

\usepackage{mathtools}
\usepackage{csquotes}

\DeclarePairedDelimiter\autobracket{(}{)}
\newcommand{\br}[1]{\autobracket*{#1}}                  
\DeclarePairedDelimiter\sparen{[}{]}
\newcommand{\sbr}[1]{\sparen*{#1}}                      
\DeclarePairedDelimiter\autobracketc{\lbrace}{\rbrace}
\newcommand{\cbr}[1]{\autobracketc*{#1}}                
\DeclarePairedDelimiter\anglegrackets{\langle}{\rangle} 

\newtheorem{example}{Example}

\newcommand{\eps}{\varepsilon}
\renewcommand{\th}{\theta}

\newcommand{\E}{\mathbb{E}}
\newcommand{\N}{\mathcal{N}}
\newcommand{\KL}{\text{KL}}
\newcommand{\LB}{\mathcal{L}}
\newcommand{\brt}{\br{\th}}
\newcommand{\lam}{\lambda}

\newcommand{\I}{\mathcal{I}}
\newcommand{\natgrad}{\tilde{\nabla}}
\newcommand{\norm}[1]{\vert\vert{#1}\vert\vert}
\renewcommand{\S}{\Sigma}
\newcommand{\iS}{\Sigma^{-1}}
\newcommand{\diag}[1]{\text{diag}\br{{#1}}}

\newcommand{\eq}[1]{eq.~\eqref{#1}}
\newcommand{\Ch}[1]{Ch.~{#1}}

\newcommand{\abr}[1]{\anglegrackets*{#1}}
\newcommand{\M}{\mathcal{M}}
\newcommand{\R}{\mathbb{R}}
\newcommand{\grad}{\text{grad}}
\newcommand{\proj}{\text{Proj}}
\newcommand{\iI}{\mathcal{I}^{-1}}
\newcommand{\brz}{\br{\zeta}}
\renewcommand{\vec}{\text{vec}}
\newcommand{\D}{\mathcal{D}}
\newcommand{\lik}{p\br{y\vert \theta}}

\renewcommand{\d}{\text{d}}               
\newcommand{\Eq}[1]{\E_{#1}}

\DeclareMathOperator*{\argmin}{arg\,min}
\DeclareMathOperator*{\argmax}{arg\,max}

\newcommand{\bx}{\bm{x}}                
\newcommand{\by}{\bm{y}}                
\newcommand{\bth}{{\bm{\theta}}}        
\newcommand{\brbth}{\br{\bm{\theta}}}   
\newcommand{\bz}{{\bm{\zeta}}}          
\newcommand{\bl}{{\bm{\lambda}}}        
\newcommand{\q}{q\brbth}                
\newcommand{\bmu}{{\bm{\mu}}}           
\newcommand{\bxi}{{\bm{\xi}}}           

\usepackage{glossaries}
\newacronym{slfn}{SLFN}{Single Layer Feedforward Network}
\newacronym{ann}{ANN}{Artificial Neural Network}
\newacronym{bnn}{BNN}{Bayesian Neural Network}
\newacronym{mc}{MC}{Monte-Carlo}
\newacronym{vi}{VI}{Variational Inference}
\newacronym{ml}{ML}{Machine Learning}
\newacronym{mcmc}{MCMC}{Monte Carlo Markov Chain}
\newacronym{dl}{DL}{Deep Learning}
\newacronym{spd}{SPD}{Symmetric and Positive Definite}
\newacronym{rsgd}{RSGD}{Riemann Stochastic Gradient Descent}
\newacronym{sgd}{SGD}{Stochastic Gradient Descent}
\newacronym{von}{VON}{Variational Online Newton}
\newacronym{vogn}{VOGN}{Variational Online Gauss-Newton}
\newacronym{adam}{ADAM}{Adaptive Moment Estimation}
\newacronym{rmsprop}{RMSProp}{Root Mean Squared Propagation}
\newacronym{mgvb}{MGVB}{Manifold Gaussian Variational Bayes}
\newacronym{emgvb}{EMGVB}{Exact Manifold Gaussian Variational Bayes}
\newacronym{qbvi}{QBVI}{Quasi Black-Box Variational Inference}
\newacronym{ngvi}{NGVI}{Natural-Gradient Variational Inference}
\newacronym{ngbbvi}{NG-BBVI}{Natural-Gradient Black-Box Variational Inference}
\newacronym{vadam}{VADAM}{Variational ADAM}
\newacronym{bbvi}{BBVI}{Black-Box Variational Inference}
\newacronym{bbb}{BBB}{Bayes-By-Backprop}
\newacronym{ffvi}{FFVI}{Fixed-Form Variational Inference}
\newacronym{kl}{KL}{Kullback-Leibler}
\newacronym{lb}{LB}{Lower Bound}
\newacronym{fim}{FIM}{Fisher Information Matrix}
\newacronym{mh}{MH}{Metropolis-Hastings}
\newacronym{hmc}{HMC}{Hamiltonian Monte Carlo}
\newacronym{gp}{GP}{Gaussian Process}
\newacronym{mcd}{MCD}{Monte Carlo Dropout}

\makeglossaries
\newcommand{\SectionNameWithGlossary}[1]{\texorpdfstring{\acrfull{#1}}{#1}}
\usepackage{multirow}

\title{Bayesian Learning for Neural Networks:\\ an algorithmic survey}

\author{%
  Martin Magris and Alexandros Iosifidis\\
  DIGIT, Department of Electrical and Computer Engineering, Aarhus University, Denmark\\
  \texttt{\{magris,ai\}@ece.au.dk} \\
}

\begin{document}

\maketitle
\begin{abstract}
The last decade witnessed a growing interest in Bayesian learning. Yet, the technicality of the topic and the multitude of ingredients involved therein, besides the complexity of turning theory into practical implementations, limit the use of the Bayesian learning paradigm, preventing its widespread adoption across different fields and applications. This self-contained survey engages and introduces readers to the principles and algorithms of Bayesian Learning for Neural Networks. It provides an introduction to the topic from an accessible, practical-algorithmic perspective. Upon providing a general introduction to Bayesian Neural Networks, we discuss and present both standard and recent approaches for Bayesian inference, with an emphasis on solutions relying on Variational Inference and the use of Natural gradients. We also discuss the use of manifold optimization as a state-of-the-art approach to Bayesian learning. We examine the characteristic properties of all the discussed methods, and provide pseudo-codes for their implementation, paying attention to practical aspects, such as the computation of the gradients. 
\end{abstract}

\section{Introduction}

\acrfull{ml} techniques have been proven to be successful in many prediction and classification tasks across natural language processing  \citep{young_recent_2018}, computer vision  \citep{krizhevsky_imagenet_2017}, time-series  \citep{langkvist2014review} and finance applications  \citep{dixon2020machine}, among the several others.
The widespread of \acrshort{ml} methods in diverse domains is found due to their ability to scale and adapt to data, and their flexibility in addressing a variety of problems while retaining high predictive ability. Recently, Bayesian methods have gained considerable interest in \acrshort{ml} as an attractive alternative to the classical methods providing point estimations for their inputs. Despite the numerous advantages that traditional \acrshort{ml} methods offer, they are, broadly speaking, prone to overfitting, dimming their generalization capabilities and performance on unseen data. Furthermore, an implicit consequence of the classical point estimation and modeling setup is that it delivers models that are generally incapable of addressing uncertainties. This inability is twofold, as it includes both the estimation and prediction aspects. Indeed, as opposed to the typical practice of statistical modeling and, e.g., econometrics methods, \acrshort{ml} methods do not directly tackle aspects related to the significance and uncertainties associated with the estimated parameters. At the same time, predictions correspond to simple point estimates without reference to the confidence levels that such estimates have. Whereas some models have been developed to, e.g., provide confidence intervals over the forecasts \citep[e.g.][]{gal2016dropout}, it has been observed that such models are generally overconfident. To estimate uncertainties implicitly embedded in \acrshort{ml} models, Bayesian inference provides an immediate remedy and stands out as the main approach.

Bayesian methods have gained considerable interest as an attractive alternative to point estimation, especially for their ability to address uncertainty via posterior distribution, generalize while reducing overfitting  \citep{hoeting1999bayesian}, and for enabling sequential learning  \citep{freitas2000hierarchical} while retaining prior and past knowledge. Although Bayesian principles have been proposed in \acrshort{ml} decades ago  \citep[e.g.][]{mackay1992bayesian,mackay1995probable,lampinen2001bayesian}, it has been only recently that fast and feasible methods boosted a growing use of Bayesian methods in complex models, such as deep neural networks  \citep{osawa2019practical,khan2018fastscalable,khan2018fastyetsimple}. 

The most challenging task in following the Bayesian paradigm is the computation of the posterior. In the typical \acrshort{ml} setting characterized by a high number of parameters and a considerable size of data, traditional sampling methods are prohibitive, and alternative estimation approaches such as \acrfull{vi} have been shown to be suitable and successful  \citep{saul1996mean,wainwright2008graphical,hoffman2013stochastic,blei2017variational}. Furthermore, recent research advocates the use of natural gradients for boosting the optimum search and the training  \citep{wierstra2014natural}, enabling fast and accurate Bayesian learning algorithms that are scalable and versatile. 

Recent years witnessed enormous growth in the interest related to Bayesian \acrshort{ml} methodologies and several contributions in the field. This survey aims at summarizing the major methodologies nowadays available, presenting them from an algorithmic, empirically-oriented perspective. With this rationale, this paper aims to provide the reader with the basic tools and concepts to understand the theory behind Bayesian \acrfull{dl} and walk through the implementation of the several Bayesian estimation methodologies available.
We should note that the focus of this paper is purely on Bayesian methods. Indeed there are a number of network architectures that can resemble a Bayesian framework by, e.g., creating a distribution for the outputs, e.g., Deep Ensembles \citep{osband2018randomized}, Batch Ensembles \citep{wen2020batchensemble}, Layer Ensembles \citep{oleksiienko2022layer}, or Variational Neural Networks \citep{oleksiienko2022variational}. These solutions, based on particular network designs, are, however, not implicitly Bayesian and out of scope in our context.
Other surveys and tutorials do exist on the general topic \citep[e.g.][along with several lecture notes available online]{jospin2022hands,Heckerman2008}, yet the focus of this paper is on algorithms and mainly devoted to \acrfull{vi} methods. In fact, despite the wide number of \acrshort{vi} and non-\acrshort{vi} methods published in the last decade, a comprehensive survey embracing and discussing all of them (or perhaps the major ones) is missing, and non-experts will easily find themselves lost in their pursuit to comprehend and different notions and processing steps in different methodologies. By filling this gap, we aim to promote applications and research in this area.

\subsection{The Bayesian paradigm}
The Bayesian paradigm in statistics is often opposed to the pure frequentist paradigm, a major area of distinction being in hypothesis testing \citep{etz2018become}. The Bayesian paradigm is based on two simple ideas. The first is that probability is a measure of belief in the occurrence of events, rather than just some limit in the frequency of occurrence when the number of samples goes toward infinity. The second is that prior beliefs influence posterior beliefs  \citep{jospin2022hands}.
The above two are summarized in the Bayes theorem, which we now review.
Let $\D$ denote the data and $p\br{\D\vert \bth}$ the likelihood of the data based on a postulated model with $\bth \in \Theta$ a $k$-dimensional vector of model parameters. Let $p\brbth$ be the prior distribution on $\bth$. The goal of Bayesian inference is the estimation of the posterior distribution \citep[e.g.,][]{gelman1995bayesian}
\begin{equation} \label{eq:bayes_th}
p\br{\bth\vert \D} = \frac{p\br{\D,\bth}}{p\br{\D}} = \frac{p\brbth p\br{\D\vert \bth}}{p\br{\D}}, 
\end{equation}
where $p\br{\D}$ is referred to as evidence or marginal likelihood, since $p\br{\D}=\int_\Theta p\brbth p\br{\D\vert \bth} d\bth$. $p\br{\D}$ acts as a normalization constant for retrieving an actual probability distribution for $p\br{\bth\vert \D}$.
In this light, as opposed to the frequentist approach, it becomes clear that the unknown parameter $\bth$ is treated as a random variable. The prior probability $p\brbth$, which intuitively expresses in probabilistic terms any knowledge about the parameter before the data has been collected, is updated in the posterior probability $p\br{\bth\vert \D}$, mixturing prior knowledge and evidence supported by the data through the model's likelihood. 
Bayesian inference is generally difficult due to the fact that the marginal likelihood is often intractable and of unknown form. Indeed, only for a limited class of models, where the prior is so-called conjugate to the likelihood, the calculation of the posterior is analytically tractable. Standard examples are Normal likelihoods and prior (resulting in Normal posteriors) or Poisson likelihoods with Gamma priors (resulting in Negative Binomial posteriors). Yet, already for the simple linear regression example, Bayesian derivation is rather tedious, and already for the logistic regression, closed-form solutions are not generally available.
It is clear that in complex models, such as deep neural networks typically used in \acrshort{ml} applications, Bayesian inference can be tackled neither analytically nor numerically (consider that the integral in the marginal likelihood is multivariate, over as many dimensions as the number of parameters). 

\acrfull{mc} methods for sampling the posterior are certainly a possibility that has been early explored and adopted. While it still remains a valid and appropriate method for performing Bayesian inference in retractable settings,  especially in high-dimensional applications, the \acrshort{mc} approach is challenging and may become infeasible, mainly because of the need for an implicit high-dimensional sampling scheme, which is generally time-consuming and computationally demanding. 
As an alternative approach, \acrshort{vi} gained much attention in recent years. \acrshort{vi} turns the integration Bayesian problem in \eq{eq:bayes_th} into an optimization problem. The idea behind \acrshort{vi} is that of targeting an approximate form of the posterior distribution, perhaps chosen within a family of well-behaved distributions, and finding the corresponding parameter that optimizes a specific objective, i.e., that is optimal under some criterion. 

In the following subsection, we review the standard non-Bayesian approach
for neural network parameter estimation (Section \ref{subsec:standardnn}), we introduce Bayesian Neural Networks (BNNs) (Section \ref{subsec:bnn}), and we provide some motivation for their use, also recalling some literature about their applications (Section \ref{subsec:motivation_BNNs}). After providing the reader an introduction to standard and Bayesian neural networks, we introduce \acrshort{vi} in Section \ref{subsec:vi}, we describe the standard framework used in Bayesian learning, and we discuss how the standard \acrfull{sgd} approach can be used for solving the optimization problem therein (Section \ref{subsec:vi:sgd}).

\subsection{Standard and Bayesian Neural Networks}\label{sec:bnn}
A \acrfull{bnn} is an \acrfull{ann} trained with Bayesian Inference \citep{jospin2022hands}. In the following, we provide a quick overview of \acrshort{ann}s and their typical estimation based on Backpropagation (Section \ref{subsec:standardnn}). We then describe what a \acrfull{bnn} is (Section \ref{subsec:bnn}), provide motivations on why to use a \acrshort{bnn}, over a standard \acrshort{ann} (Section \ref{subsec:motivation_BNNs}), and lastly introduce \acrfull{vi} (Section \ref{subsec:vi}).

\subsubsection{Artificial Neural Networks}\label{subsec:standardnn}

For completeness, we review the general ingredients, principles, ideas, and standard terminology behind \acrfull{ann}. A comprehensive and more detailed introduction to the topic is here out of scope; the interested reader can e.g., consult the accessible book \citep{haykin1998neural}.

\begin{figure}[!h]
    \centering
    \includegraphics[scale=1]{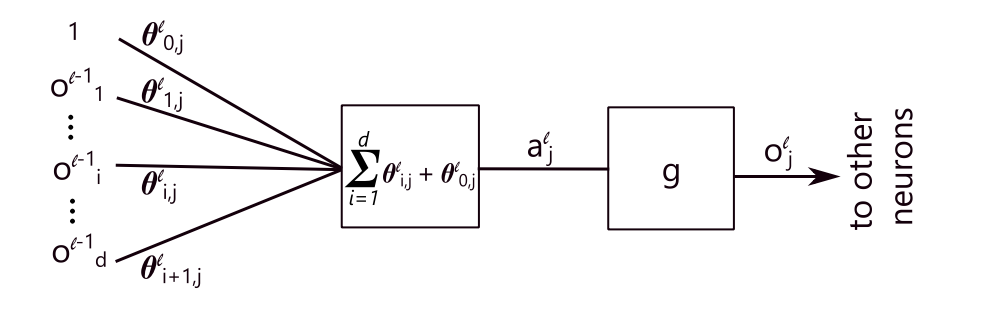}
    \caption{Representation of the operations within the $j$-th neuron at layer is $l$. Connections between this neuron and neurons in layer $l-1$  are represented by lines corresponding to weights $\th^l_{\cdot j}$. The inputs to the neuron $o^{l-1}_{\cdot}$ interact with the weights $\th^l_{\cdot j}$, computing the weighted sum $a^l_j$. The so-called \textit{activation function} $g(\cdot)$ is applied to $a^l_j$ leading to the output $o^l_j$, which is sent to nodes at layer $l+1$.} 
    \label{fig:neuron_example}
\end{figure}

\textit{Neurons} are elementary building blocks which can be thought of as processing units that, when combined, constitute a neural network. Each neuron processes the information presented to its input by applying a transformation to it. When affine neurons are used, the transformation corresponds to computing the weighted sum of the inputs to the neuron (received from the neurons that are connected to it or corresponding to the inputs to the neural network) and generates a value, which is further introduced to a (usually nonlinear) \textit{activation function} to produce the neuron's output (input to other neurons or the neural network output). 
In order to account for the need of a shift to the value needed to produce an activation response, a bias is also added as an input to the activation function, which is commonly included in the weighted sum by augmenting the input to the neuron with an additional input with a constant value of 1, associated with the corresponding bias term. 
While activation functions squeezing their outputs to a pre-determined range of values, like the sigmoid (with outputs in $[0,1]$) or the tanh (with outputs in $[-1,1]$) functions, have been widely used in the past, piece-wise linear functions, like the Rectified Linear Unit (ReLU) or the parametric ReLU functions \citep{he2015delving}, are nowadays widely adopted in building the hidden layers of neural networks. Linear and softmax activation functions are commonly used in the output layer for regression and classification problems, respectively. A common characteristic of activation functions used in neural networks is that they are \textit{differentiable} with respect to their parameters over the range of their inputs.
The transformation performed by an affine neuron is illustrated in Figure \ref{fig:neuron_example}.   

\begin{figure}
    \centering
    \includegraphics[scale=0.8]{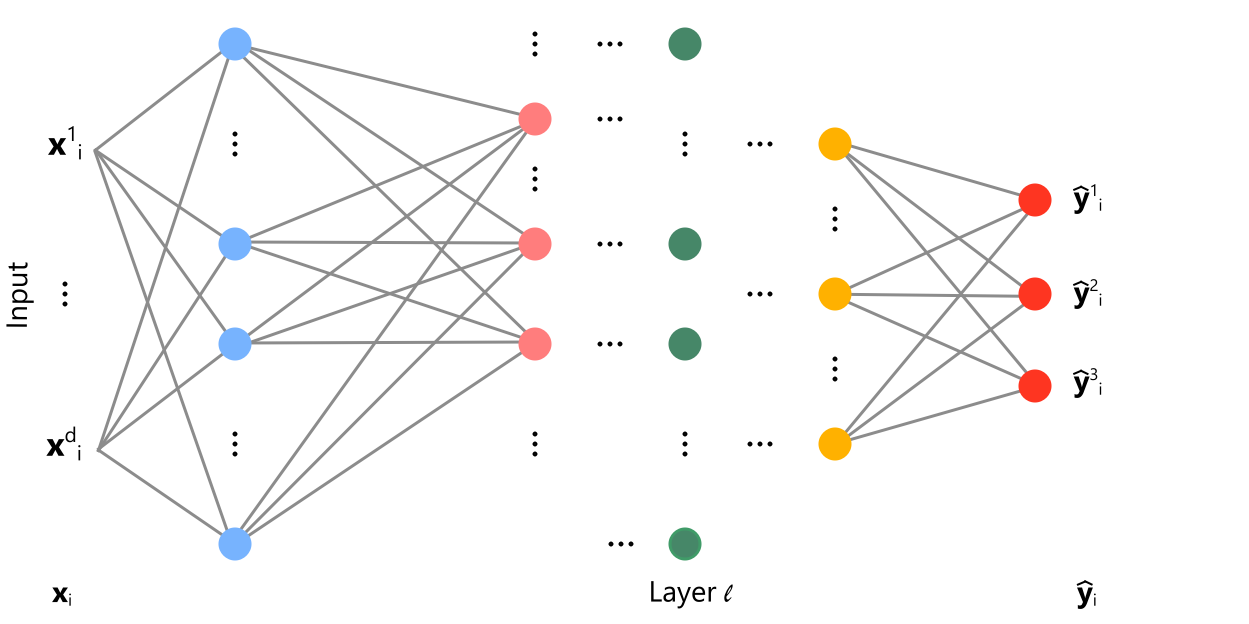}
    \caption{A feedforward network with multiple layers. Dots represent neurons across different layers (colors). The $d$-dimensional input vector $\bx_i = [x_i^1,\dots,x_i^d]^T$ is sequentially parsed to the output, from left to right, following the connections represented in grey which correspond to the weights of the network's layers.} 
    \label{fig:network_example}
\end{figure}

Whenever the information flow between neurons has no feedback (i.e., neurons do not process information resulting from their outputs), in the sense that information flows from the input through the neurons producing the output of the network, the network is referred to as \textit{feedforward}. Neurons are arranged in layers, and a network formed by neurons in one layer is called \textit{single layer network}. When more than one layer forms a neural network, layers are generally called \textit{hidden layers} since they stand between the input and the output, i.e., the \enquote{tangible} information, which consists of the input samples and their classification targets/outputs. 
A feedforward neural network receiving as input a $d$-dimensional vector and producing a $3$-dimensional output is shown in Figure \ref{fig:network_example}.

The most relevant feature of a neural network is its capacity of \textit{learning}. This corresponds to the ability to improve its outputs (performance in classification) by tuning the parameters (weights and biases) of its neurons. Learning algorithms of neural networks use a set of training data to iteratively update the parameters of a neural network such that some error measure is decreased or some performance measure is increased \citep[see, e.g.,][]{Goodfellow-et-al-2016}.
The data $\D$ consists of vectors $\D_i = \cbr{\by_i,\bx_i}$, with $\bx_i$ representing an input and $\by_i$ the corresponding target for $i=1,...,N$. Let $\hat{\by}_i$ denote the output of the network corresponding to the sample $\bx_i$, that is $\hat{\by}_i = \text{NN}_\bth\br{\bx_i}$, with $\text{NN}_\bth\br{\bx_i}$ denoting a Neural Network parametrized over $\bth$ and evaluated at $\bx_i$. An error function $E\br{\D,\bth}$ is defined at a particular parameter $\bth$, which is used to guide the learning process. Several error functions have been used to this end, the most widely adopted ones being the mean-squared error (suitable for regression problems) and the cross-entropy (suitable for classification problems). The \textit{gradient} of the error between the network's outputs $\hat{\by}_i$ and the targets $\by_i$ over the entire data set (full-batch) or a subset of the data (mini-batch) is commonly used to update the network parameter values through an iterative optimization process, commonly a variant of the Backpropagation algorithm \citep{rumelhart1986learning}. Widely used iterative optimization methods are the \acrfull{sgd} \citep{robbins1951stochastic}, \acrfull{rmsprop} \citep{tieleman2012lecture} and \acrfull{adam} \citep{kingma2014adam}.

While feedforward neural networks with affine neurons have been briefly described above, a large variety of neural networks have been proposed and used for modeling different input-output data relationships. Such networks follow the main principles as those described above (i.e., they are formed by layers of neurons, which perform transformations followed by differentiable activation functions), but they are realized by using different types of neurons and/or transformations. 
Examples include the Radial Basis Function (RBF) networks \citep{broomhead1988multivariable}, which replace affine transformations with distance-based transformations, Convolutional Neural Networks \citep{toshiteru1988artificial}, which receive a tensor input and use neurons performing convolution, Recurrent Neural Networks (e.g., Long-Short Term Memory (LSTM) \citep{hochreiter1997long} and Gated Recurrent Unit (GRU) \citep{kyunghyun2014properties} networks), which model sequences of their inputs by using recurrent units, and specialized types of neural networks, such as the Temporal-Augmented Bilinear Layer (TABL) network \citep{tran_temporal_2019} based on bilinear mapping, and the Neural Bag-of-Features network \citep{passalis2020temporal}, extending the classical Bag-of-Features model with a differentiable processing suitable to be used in combination with other types of neural network layers.

\subsection{Motivation for adopting Bayesian Neural Networks}\label{subsec:motivation_BNNs}
Bayesian Neural Networks are interesting tools under three perspectives: (i) theoretical, (ii) methodological, and (iii) practical. In the following, we shall briefly discuss what we mean by the above three interconnected perspectives.

From a theoretical perspective, \acrshort{bnn}s allow for differentiating and quantifying two different sources of uncertainty, namely epistemic uncertainty, and aleatoric uncertainty \citep[see, e.g.][from a \acrshort{ml} perspective]{der2009aleatory}. Epistemic uncertainty is the one referring to the lack of knowledge, and it is captured by $p\br{\bth\vert \D}$. In light of the Bayes theorem, epistemic uncertainty can be reduced with the use of additional data so that the lack of knowledge is addressed as more data are collected. After the data is collected, this results in the update of the prior belief (before the experiment is conducted) to the posterior. Thus, the Bayesian perspective allows the mixing of expert knowledge with experimental evidence. This is quite relevant in small-sample applications where the amount of collected data is inappropriate for classical statistical tools and results to apply (e.g., inference based on asymptotic theory), yet it nevertheless allows the update of the a-priori belief on the parameters, $p\brbth$, into the posterior. On the other hand, the likelihood term captures the aleatoric uncertainty, that is the intrinsic uncertainty naturally embedded in the data, i.e., $\lik$, in the Bayesian framework is clearly distinguished and separated from the aleatoric one.

Methodologically, is remarkable the ability of Bayesian methods to learn from small data and eventually converge to, e.g., non-Bayesian maximum likelihood estimates or, more generally, to agree with alternative frequentist methods. When the amount of the collected data overwhelms the role of the prior in the likelihood-prior mixture, Bayesian methods can be clearly seen as generalizations of standard non-Bayesian approaches. Within the Bayesian methods family, certain research areas such as PAC-Bayes \citep{alquier2021user}, Empirical Bayes \citep{casella1985introduction} and Approximate Bayes Computations \citep{csillery2010approximate} deal with such connections very tightly. In this regard, there are many examples in the statistics literature; we focus on the \acrshort{ml} perspective. For instance, regularization, ensemble, meta-learning, Monte-Carlo dropout, etc., can all be understood as Bayesian methods, and, e.g., Variational Bayes can be seen as standard linear regression \citep{salimans2013fixed}. More in general, many \acrshort{ml} methods can be seen as approximate Bayesian methods, whose approximate nature makes them simpler and of practical use. Furthermore, as the learned posterior can be reused and re-updated once new data become available, Bayesian learning methods are well-suited for online learning \citep{opper1999bayesian}. In this regard, also the explicit use of the prior in Bayesian formulations is aligned with the No-Free-Lunch Theorem \citep{wolpert1996lack} whose philosophical interpretation, among the others, is that any supervised algorithm implicitly embeds and encodes some form of prior, establishing a tight connection with Bayesian theory \citep{serafino2013no,guedj2021still}.

From a practical perspective, the Bayesian approach implicitly allows for dealing with uncertainties, both in the estimated parameters and in the predictions. For a practitioner, this is by far the most relevant aspect in shifting from a standard \acrshort{ann} approach to \acrshort{bnn}s. Thus, with little surprise, Bayesian methods have been well-received in high-risk application domains where quantifying uncertainties is of high importance. Examples can be found across different fields, such as industrial applications \citep{vehtari1999bayesian}, medical applications \citep[e.g.][]{chakraborty2012applications,kwon2020uncertainty,lisboa2003bayesian}, finance \citep[e.g.][]{jang2017empirical,sariev2020bayesian,magris2022bayesian,magris2022exact}, fraud detection \citep[e.g.][]{viaene2005auto}, engineering \citep[e.g.][]{cai2018vibnn,du2020predicting,goh2005bayesian}, and genetics \citep[e.g.][]{ma1999biological,liang2004hierarchical,waldmann2018approximate}.

As widely recognized, the estimation of \acrshort{bnn} is not a simple task due to the generally non-conjugacy between the prior and the likelihood and the non-trivial computation of the integral involved in the marginal likelihood. For this reason, application of \acrshort{bnn}s is relatively infrequent, and their use is not widespread across the different domains. As of now, applying Bayesian principles in a plug-and-play fashion is challenging for the general practitioner. On top of that,  several estimation approaches have been developed, and navigating through them can indeed be confusing. In this survey, we collect and present parameter estimation and inference methods for Bayesian \acrshort{dl} at an accessible level to promote the use of the Bayesian framework.

\subsection{Bayesian Neural Networks}\label{subsec:bnn}
From the description in Section \ref{subsec:standardnn}, it can be seen that the goal of approximating a function relating the input to the output in classical \acrshort{ann}s is treated under an entirely deterministic perspective. Switching towards a Bayesian perspective in mathematical terms is rather straightforward. In place of estimating the parameter vector $\bth$, \acrshort{bnn}s target the estimation of the posterior distribution $p\br{\bth|\D_x,\D_y}$, that is \citep{jospin2022hands}:
\begin{equation}\label{eq:bayes_th_BNN}
p\br{\bth|\D_x,\D_y} =\frac{p\br{\D_y\vert \D_x,\bth}p\br{\bth}}{\int_\Theta p\br{\D_y\vert \D_x,\bth'}p\br{\bth'}d\bth'}\text{,}
\end{equation}
which stands as a simple application of the Bayes theorem.
Here we assume, as it is usually the case, that the data $\D$ is composed of an input set $\D_x$ and the corresponding set of outputs $\D_y$. In general, $\D_x$ is a matrix of regressors, and $\D_y$ is either the vector or matrix (depending on whether the nature of the output is univariate or not) of the variables that the networks aim at modeling based on $\D_x$. Alternatively but analogously, $\D$ can be thought as the collection of all input-output pairs $\D = \cbr{\bm{y}_i,\bm{x}_i}_{i=1}^N$, where $N$ denotes the sample size, and $\bm{x}_i$ and $\bm{y}_i$ are the input and output vectors of observations for the $i$-th sample, respectively. Using this notation, $\D_x=\cbr{\bm{x}_i}_{i=1}^N$ and $\D_y=\cbr{\bm{y}_i}_{i=1}^N$.

While \eq{eq:bayes_th_BNN} provides a theoretical prescription for obtaining the posterior distribution, in practice solving for the form of the posterior distribution and retrieving its parameters is a very challenging task.
The estimation of a \acrshort{bnn} with \acrshort{mc} techniques and \acrshort{vi} is discussed in the remainder of the review, here we continue the discussion towards different aspects.

Equation (\ref{eq:bayes_th_BNN}) involves all the ingredients required for performing Bayesian inference on \acrshort{ml} models, and specifically neural networks. 
In the first place, \eq{eq:bayes_th_BNN} involves a likelihood function for the data $\D_y$ conditional on the observed sample $\D_x$ and the parameter vector $\bth$. The forward pass parses the input into predictions via some parameter values, such outputs (conditional on the data and the parameters) follow a prescribed likelihood function. Intuitive examples are the Gaussian likelihood (for regression) and the Binomial one (for classification). An underlying neural network is implicit in the likelihood term $p\br{\D_y\vert \D_x,\bth}$, that links the inputs to the outputs.
In other words, as is the case for \acrshort{ann}s, the first step in designing a \acrshort{bnn} is that of identifying a suitable neural network architecture (e.g., how many layers and of which kind and size) followed by a reasonable assumption for the likelihood function. 

A major difference between \acrshort{ann}s and \acrshort{bnn}s is that the latter requires the introduction of the prior distribution over the model parameters. After all, a prior must be in place for Bayesian inference to be performed; thus, priors are required in the \acrshort{bnn} setup \citep{jospin2022hands}. This means that the practitioner needs to decide on the parametric form of the prior over the parameters.

\begin{example}
Consider a \acrshort{bnn} to model the variables $\D_y = \cbr{y_i}_{i=1}^N$ where $y_i \in \cbr{0,1}$, based on the matrix of covariates $D_x$. The likelihood is of a certain form and parametrized over a neural network whose weights are denoted by $\bth$, i.e., $\text{NN}_\bth\br{\cdot}$. 

We can approach the above problem as a 2-class classification with $y_i \in \sbr{0,1}$, and derive the likelihood from the Bernoulli distribution
\begin{equation}
p\br{D_y\vert \D_x,\bth} = \prod_{i=1}^N \hat{p}_i^{y_i}\br{1-\hat{p}_i}^{1-y_i}\text{,}    
\end{equation}
where $\hat{p}_i = \text{NN}_\bth\br{\bm{x}_i}$ denotes the output of the network for the $i$-th sample, that is the probability that sample $i$ belongs to class 1.
The prior (on the network parameters) can be a diagonal Gaussian $p\br{\bth}=\mathcal{N}\br{\bth \vert 0, \tau I}$, where $\tau>0$ is a scalar and $I$ the identity matrix.

We can also approach the above problem as a regression to $\bm{y}_i \in \mathbb{R}^d$ and derive the likelihood from the Multivariate Normal distribution
\begin{align}
p\br{D_y\vert D_x,\bth} = &\br{2\pi}^{-Nk/2}\vert \text{det}\br{\S}\vert^{-N/2}\nonumber \\
&\times \exp\br{-\frac{1}{2} \sum_{i=1}^N\br{\bm{y}_i-\hat{\bm{y}}_i}^\top \S^{-1}\br{\bm{y}_i-\hat{\bm{y}}_i}}
\end{align}
where
$\hat{\bm{y}}_i = \text{NN}_\bth\br{\bm{x}_i}$. Assuming that the covariance matrix $\S^{-1}$ is known, the prior on $\bth$ could be as well a diagonal Gaussian. If $\S$ is unknown, the prior could be the product of the above Gaussian prior with, e.g., an Inverse Wishart prior distribution on $\S$. In this case, the goal of the Bayesian inference is that of estimating the joint posterior of $\br{\bth,\S}$.
\end{example}

The inference goal is the posterior distribution. (i) If the problem has a form for which the posterior can be solved analytically, we find $p\br{\bth\vert \D_x\D_y}$ to be of a known parametric form and identify the parameters characterizing it (standard Bayesian setting, so-called conjugacy between the prior and the likelihood, \citep[e.g.,][]{gelman1995bayesian}). (ii) In general, we may proceed via \acrshort{mc} sampling, in which case the estimation leads to a sample, of arbitrary size, approximating the true posterior. The true posterior remains unknown in its exact form, yet \acrshort{mc} enables sampling from it and thus estimating an approximate representation \citep[e.g.,][]{gamerman2006markov}, see Section \ref{sec:MC}. (iii) Alternatively, following \acrshort{vi}, one sets a certain chosen parametric form for the posterior and optimizes its parameters for a certain objective function \citep[e.g.,][]{nakajima_watanabe_sugiyama_2019}, see Section \ref{subsec:vi}. While the actual posterior remains unknown, in \acrshort{vi} one seeks an approximation that is optimal in some sense of optimization of a certain objective on the provided data.

\begin{figure}
    \centering
    \includegraphics[scale=0.9]{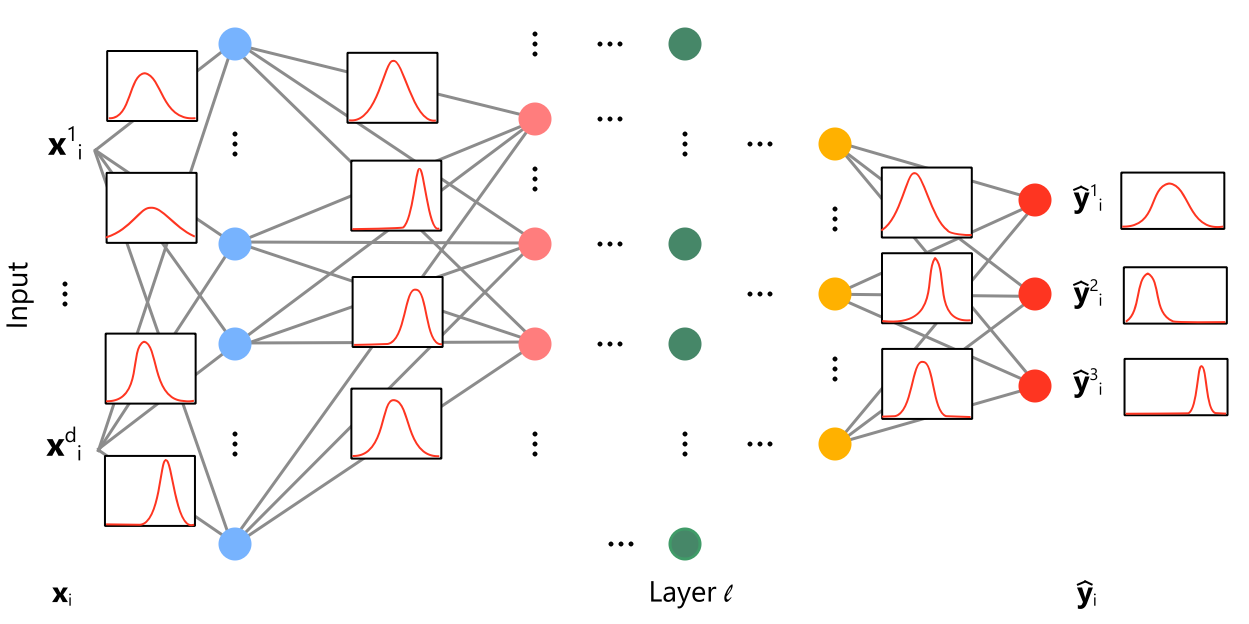}
    \caption{A \acrshort{bnn} with multiple layers. Connections correspond to random variables, and outputs here correspond to a tri-variate distribution, whose marginals are represented in the rightmost boxes.}
    \label{fig:bnetwork_example}
\end{figure}

Figure \ref{fig:bnetwork_example} provides an analogous representation of Figure \ref{fig:network_example}, now for a \acrshort{bnn}. Opposed to traditional \acrshort{ann}s, weights in \acrshort{bnn}s are stochastic and represented with distributions. A probability distribution over the weights is learned by updating the prior with the evidence supported by the data. Even though Figure \ref{fig:bnetwork_example} might give the opposite impression, the posterior over the weights is, in general, a truly multivariate distribution where independence among its dimensions generally does not hold.

While the above clarifies that the estimation goal is a distribution whose, e.g., variance can be indicative of the level of confidence in the estimated parameters, 
the uncertainty associated with the outputs and the generation of the model outputs themselves remains unaddressed.
The predictive distribution is defined as \citep[e.g.,][]{gelman1995bayesian}
\begin{equation}
    p\br{\by_i\vert \bx_i, \D} = \int_\Theta p\br{\by_i\vert \bx_i,\bth}p\br{\bth\vert \D}d\bth.
\end{equation}
As the posterior (\eq{eq:bayes_th_BNN}) is solved, the predictive distribution can also be recovered. Yet, in practice, it is indirectly sampled. Indeed, an intuitive \acrshort{mc}-related approach for approximating the predictive distribution is that of sampling $N_s$ values from the posterior to create $N_s$ realizations of the neural network, each based on a different parameter sample, which are used to provide predictions. This results in a collection of predictions that approximate the actual predictive distribution.
In this way, it is relatively simple to recover (approximations of) the predictive distribution from which, e.g., confidence intervals can be constructed.
A way to reduce the sample forecast to single values conveying relevant information is by, e.g., using common (sampling) moment estimators \citep[e.g.,][Ch. 7.2.1]{casella2021statistical}.
One may evaluate
\begin{equation}
    \hat{\by}_i = \frac{1}{N_s}\sum_{j=1}^{N_s} \text{NN}_{\bth_j}\br{\bx_i} \text{,}
\end{equation}
to approximate the posterior mean through model averaging (across the different realizations $\bth_j, \:j=1,\dots,N_s$ and thus different outputs) or compute
\begin{equation}
    \hat{\S}_{\by_i} = \frac{1}{N_s-1}\sum_{j=1}^{N_s}\bm{\varepsilon}_{j,i} \bm{\varepsilon}_{j,i}^\top \text{,} 
\end{equation}
with
\begin{equation}
    \quad \bm{\varepsilon}_{j,i} =  \text{NN}_{\bth_j}\br{\bx_i}-\hat{\by}_i\text{,}
\end{equation}
to approximate the covariance matrix, which is indicative of the uncertainty associated with the prediction. $N_s$ corresponds to the number of samples generated from the posterior and used to generate the prediction of the network $\text{NN}_{\bth_j}(\cdot)$ receiving as input $\bx_i$. In classification, one may analogously approximate predictive densities for the joint probability of the different classes and average such probabilities to summarize the average probabilities of each class and implicitly the uncertainties associated with a certain class decision, which is typically determined by the predicted class of maximum probability \citep[e.g.,][]{osawa2019practical,magris2022bayesian}:
\begin{equation}
   \hat{y}_i =\argmax_{c\, \in\, C} \hat{p}_{i,c}  \text{,}
\end{equation}
with $C$ being the total number of classes and $\hat{p}_{i,c}$ the predicted probability of class $c$ for the sample $i$.

\subsection{\SectionNameWithGlossary{vi}}\label{subsec:vi}
Let $\D$ denote the data and $p\br{\D\vert\bth}$ the likelihood of the data based on a postulated model with $\bth \in \Theta$ a $k$-dimensional vector of model parameters. Let $p\brbth$ be the prior distribution on $\bth$. The goal of Bayesian inference is the posterior distribution
\begin{equation} \label{eq:bayes_th_vi}
p\br{\bth\vert \D} = \frac{p\br{\D,\bth}}{p\br{\D}} = \frac{p\br{\D\vert \bth} p\brbth }{p\br{\D}}, 
\end{equation}
where $p\br{\D}$ is referred to as evidence or marginal likelihood, since $p\br{\D}=\int_\Theta \br{\D\vert \bth} p\brbth  d\bth$. $p\br{\bth}$ acts as a normalization constant for retrieving an actual probability distribution for $p\br{\bth\vert \D}$. Bayesian inference is generally difficult due to the fact that the evidence is often intractable and of unknown form. 
In high-dimensional applications, \acrlong{mc} methods for sampling the posterior turn challenging and infeasible, and \acrfull{vi} is an attractive alternative.

\acrshort{vi} consists in an approximate method where the posterior distribution is approximated by the so-called  {\it variational distribution} \citep[e.g.,][]{blei2017variational,nakajima_watanabe_sugiyama_2019,tran2021practical}. The variational distribution is a probability density $\q$, belonging to some tractable class of distributions $\mathcal{Q}$ such as, e.g., the Exponential family. \acrshort{vi} thus turns the Bayesian inference problem in eq. \eqref{eq:bayes_th_vi} into that of finding the best approximation $q^\star \brbth \in \mathcal{Q}$ to $p\br{\bth\vert \D}$ by minimizing the \acrfull{kl} divergence from $\q$ to $p\br{\bth\vert \D}$ \citep{kullback1951information},
\begin{equation}
    q^\star = \argmin_{q \in \mathcal{Q}} \: \KL\br{q \vert \vert p\br{\bth\vert \D}} = \argmin_{q \in \mathcal{Q}} \int \q \log \frac{\q}{p\br{\bth\vert \D}} d\bth \text{.}
\end{equation}
By simple manipulations of the \acrshort{kl} divergence definition, it can be shown that
\begin{equation}
\KL\br{q \vert \vert p\br{\bth\vert \D}} = -\int \q \log \frac{p\br{\D\vert \bth} p\brbth }{\q}d\bth + \log p\br{\D}  \text{.}
\end{equation}
Since $\log p\br{\D}$ is a constant not depending on the model parameters, the \acrshort{kl} minimization problem is equivalent to the maximization problem of the so-called \acrfull{lb} on $\log p\br{\D}$ \citep[e.g.,][]{nakajima_watanabe_sugiyama_2019}, 
\begin{equation}
    \LB\br{q} \coloneqq \int \q \log \frac{p\br{\D\vert \bth} p\brbth }{\q} d \bth
    = \Eq{q}\sbr{\log \frac{p\br{\D\vert \bth} p\brbth }{\q}} \text{.}
\end{equation}
For any random vector $\bth$ and a function $g\brbth$ we denote by $\Eq{f}[ g\brbth]$ the expectation of $g\brbth$ where $\bth$ follows a probability distribution with density $f$, i.e. $\Eq{f}[ g\brbth] =\Eq{\bth \sim f}[g\brbth]$.
To make explicit the dependence of the \acrshort{lb} on some vector of parameters $\bz$
parametrizing the variational posterior we write $\LB\br{\bz}=\LB \br{q_\bz}= \Eq{q_\bz} \sbr{\log p\brbth - \log q_\bz \brbth + p\br{\D\vert \bth}  }$.
We operate within the \acrfull{ffvi} framework, where the parametric form of the variational posterior is set \citep[e.g.,][]{tran2021practical}.
I.e., \acrshort{ffvi} seeks at finding the best $q\equiv q_\bz$ in the class $\mathcal{Q}$ of distributions indexed by a vector parameter $\bz$ that minimizes the \acrshort{lb} $\LB\br{\bz}$. In this context, $\bz$ is called {\it variational parameter}. A common choice for $\mathcal{Q}$ is the Exponential family, and $\bz$ is the corresponding natural parameter.

\subsubsection{Estimation with \SectionNameWithGlossary{sgd}}\label{subsec:vi:sgd}
A straightforward approach to maximize $\LB\br{\bz}$ is that of using a gradient-based method such as \acrfull{sgd}, \acrfull{adam} \citep{kingma2014adam}, or \acrfull{rmsprop} \citep{tieleman2012lecture}. The form of the basic \acrshort{sgd} update is
\begin{equation} \label{eq:sgd_on_zeta}
    \bz_{t+1} = \bz_t+\beta_t \left. \sbr{\hat{\nabla}_\bz\mathcal{L}\br{\bz} } \right|_{\bz = \bz_t} \text{,}
\end{equation}
where $t$ denotes the iteration, $\beta_t$ a (possibly adaptive) step size, and $\hat{\nabla}_\bz\LB\br{\bz}$ a stochastic estimate of $\nabla_\bz \LB\br{\bz}$. The derivative, considered with respect to $\bz$, is evaluated at $\bz = \bz_t$.

Under a pure Gaussian variational assumption, it is instinctive to optimize the \acrshort{lb} for the mean vector $\bz_1 = \bmu$ and variance-covariance matrix $\bz_2 = \S$. In the wider \acrshort{ffvi} setting with $\mathcal{Q}$ being the Exponential family, the \acrshort{lb} is often optimized in terms of the natural parameter $\bl$  \citep{wainwright2008graphical}. 
The application of the \acrshort{sgd} update based on the standard gradient is problematic because it ignores the information geometry of the distribution $q_\bz$  \citep{amari1998natural}, as it implicitly relies on the Euclidean distance to capture the dissimilarity between two distributions in terms of the euclidean norm $\vert \vert \bz_t - \bz \vert \vert^2$, which can be a quite poor and misleading measure of dissimilarity \citep{khan2018fastyetsimple}. By replacing the Euclidean norm with the \acrshort{kl} divergence, the \acrshort{sgd} update results in the following natural gradient update:
\begin{align}\label{eq:sgd_nat_update}
   \bl_{t+1} =\bl_t+\beta_t\sbr{\tilde{\nabla}_\bl\mathcal{L}\br{\bl}} \text{.}
\end{align}
The natural gradient update results in better step directions toward the optimum when optimizing the distribution parameter. The natural gradient of $\LB\br{\bl}$ is obtained by rescaling the gradient $\nabla_\bl \LB\br{\bl}$ by the inverse of the \acrfull{fim} $\I_\bl$,
\begin{equation} \label{eq:def_natgrad}
  \tilde{\nabla}_\bl \mathcal{L}\br{\bl} = \iI_\bl\nabla\LB_\bl\br{\bl},
\end{equation}
where subscript in $\iI_\bl$ remarks that the \acrshort{fim} is expressed in terms of the natural parameter $\bl$.
By replacing in the above $\nabla_\bl \LB\br{\bl}$ with a stochastic estimate $\hat{\nabla}_\bl\LB\br{\bl}$ one obtains a stochastic natural gradient update.

\begin{example}[Problem statement for \SectionNameWithGlossary{vi}]
Consider a \acrshort{bnn} to model the targets $\by_i$, based on the covariates $\bx_i$. The likelihood, of a certain form, is parametrized over a neural network, whose weights are denoted by $\bth$. The prior could be a Gaussian distribution with, e.g., zero-mean, diagonal $p\brbth=\mathcal{N}\br{\bth \vert \bm{0}, I/ \tau}$ or not $p\brbth=\mathcal{N}\br{\bth \vert \bm{0}, \S_0}$. $\mathcal{Q}$ is the set of multivariate Gaussian distributions, specified, e.g., in terms of the natural parameter $\bl$. 

The objective is that of finding the corresponding variational parameter such that the \acrshort{lb} $\Eq{q_\bl} \sbr{\log p\brbth - \log q_\bl \brbth + p\br{\D\vert \bth}  }$ is maximized.
The update of the variational parameter $\bl$ follows a gradient-based method with natural gradients. The training terminates after the \acrshort{lb} $\LB\br{\bl}$ does not improve for a certain number of iterations: the terminal $\bl$ provides the natural parameter of the variational posterior approximation minimizing the \acrshort{kl} divergence to the true posterior $p\br{\bth \vert \mathcal{D}}$.
\end{example}

\section{Sampling methods} \label{sec:MC}
\subsection{\SectionNameWithGlossary{mcmc}}
\acrfull{mcmc} is a set of methods for sampling from a probability distribution. \acrshort{mcmc}s have numerous applications, and especially in Bayesian statistics are a fundamental tool. The foundation of \acrshort{mcmc} methods are Markov Chains, stochastic models describing a sequence of events in which the probability of each event depends only on the state of the previous one \citep{gagniuc2017markov}. 
By constructing a Markov Chain that has the desired distribution as its stationary distribution, towards which the sequence eventually converges, one can obtain samples from it, i.e., one can sample any generic probability distribution, including, e.g., a complex, perhaps multi-modal, Bayesian posterior. Early samples may be autocorrelated and not representative of the target distribution, so that \acrshort{mcmc} methods generally require a burnout period before attaining the so-called stationary distribution. In fact, while the construction of a Markov Chain converging to the desired distribution is relatively simple, determining the number of steps to achieve such convergence with an acceptable error is much more challenging and strongly dependent on the initial setup and starting values. With burnout, the large collection of samples is practically subsampled by discarding an initial fraction of draws (e.g., 20\%) to obtain a collection of approximately independent samples from the desired distribution. An accessible introduction to Markov Chains can be found in \citep{gagniuc2017markov}, for a dedicated monograph on \acrshort{mcmc} methods oriented toward Bayesian statistics and applications see, e.g., \citep{gamerman2006markov}.

Within the class of \acrshort{mcmc} methods, some popular ones are not effective in large Bayesian problems such as \acrshort{bnn}s. For example, the plain Gibbs sampler \citep{geman1984stochastic}, despite its simplicity and desirable properties \citep{casella1992explaining}, suffers from residual autocorrelation between successive samples and becomes increasingly difficult as the dimensionality increases in multivariate distributions \citep[e.g.][\Ch{4}]{lynch2007introduction}. We review the most widespread \acrshort{mc} approaches in the context of performing Bayesian learning for Neural networks.

\subsection{\SectionNameWithGlossary{mh}}
The \acrfull{mh} algorithm \citep{metropolis1953equation,hastings1970monte} is particularly helpful in Bayesian inference as it allows drawing samples from any probability distribution $p$, given that a function $f$ proportional up to a constant to $p$ can be computed. This is particularly convenient as it allows to sample a Bayesian posterior by only evaluating $ f\br{\bth} = p\br{\bth\vert y}p\brbth$, completely excluding the normalization factor from the computations. The values of the Markov Chain are sampled iteratively, with each value depending solely on the preceding one: at each iteration, based on the current value, the algorithm picks a candidate value $\bth$ (proposed value), which is either accepted or rejected randomly with a probability that depends on the current and earlier values. Upon acceptance, the proposed value is used for the next iteration, otherwise is discarded, and the current value is used in the next iteration. As the algorithm proceeds and more sample values are generated, the sampled-value distribution more and more closely approximates the target distribution $p$.

A key ingredient in \acrshort{mh} is the proposal density determining the drawing of the proposed value at each iteration. This is formalized by an arbitrary probability density $g\br{\bth^\star\vert\cdot}$, upon which depends the probability of drawing $\bth^\star$ given the previous value $\bth$. $g$ is usually assumed symmetric, and a common choice is provided by a Gaussian distribution centered on $\bth$. Algorithm \ref{alg:MH} summarizes the above steps.

\begin{algorithm}
\caption{Metropolis-Hastings}
\label{alg:MH}
\begin{algorithmic}
\State Initialize $\bth_0$
\For{$t = 0$ to $N_\text{max}$}
    \State Simulate $x^\star \sim g \br{\bth^\star\vert \bth_t}$
    \State Calculate $r = \frac{f\br{\bth^\star} q\br{\bth_t\vert \bth^\star}}{f\br{\bth_t}q\br{\bth^\star \vert \bth_{t}}}$
    \State Simulate $\alpha \sim \mathcal{U}\sbr{0,1}$
    \If{$\alpha < r$}
        \State $\bth_{t+1} = \bth^\star$
    \Else
        \State $\bth_{t+1} = \bth_t$
\EndIf
\EndFor
\State \Return $\left\lbrace\bth_t\right\rbrace_{t=B}^{N_\text{max}}$
\end{algorithmic}
\end{algorithm}

The acceptance ratio $\alpha$ is representative of the likelihood of the proposed sample $\bth^\star$ over the current one $\bth_t$ according to $p$. Indeed, $\alpha = f\br{\bth^\star}/f\br{\bth_t} = p\br{\bth^\star}/p\br{\bth_t}$ as $f$ is proportional to $p$. A proposed sample value $\bth^\star$ that is more probable than $\bth_t$ ($\alpha >1$) is always accepted; otherwise, it may be rejected with probability $\alpha$. The algorithm thus moves around the sample space, tending to stay in regions where $p$ is of high density and, seldomly, in regions of low density. The final collection of samples follows the distribution $p$. As the Markov chain eventually converges to the target distribution $p$, initial samples may be quite incompatible with $p$, especially if the algorithm is initialized at a low-density region. Thus, it is customary to discard a number $B$ of samples and retain only the subsample $\left\lbrace\bth_t\right\rbrace_{t=B}^N$. Note that by construction, successive samples of the Markov chain are correlated. Even though the chain eventually converges to $p$ nearby samples are correlated, causing a reduction of the effective sample size (e.g., for $\E_{\bth \sim p_\bth}\sbr{\bth}$ the central limit theorem applies but, e.g., the limiting variance is inflated by the non-zero autocorrelation in the chain).

An important feature of the \acrshort{mh} algorithm is that it is applicable to high dimensions as it does not suffer from the course of dimensionality problem, causing an increasing rejection rate as the number of dimensions increases. This makes \acrshort{mh} suitable for large Bayesian inference problems such as training \acrshort{bnn}s.

\subsection{\SectionNameWithGlossary{hmc}}
\acrfull{hmc} generates efficient transitions by using the derivatives of the density function being sampled by using approximate Hamiltonian dynamics, later corrected for performing an \acrshort{mh}-like acceptance step \citep{neal2011mcmc}.

HMC augments the target probability density $p\brbth$ by introducing an auxiliary momentum variable $\rho$ and generating draws from 
\begin{equation}
p\br{\rho,\bth} = p\br{\rho\vert \bth}p\brbth\text{.}   
\end{equation}
Typically the auxiliary density is taken as a multivariate Gaussian distribution, independent of $\bth$:
\begin{equation}
\rho \sim \N\br{\bm{0},\S}\text{.}
\end{equation}
$\S$ can be conveniently set to the identity matrix, restricted to a diagonal matrix, or estimated from warm-up draws. The Hamiltonian is defined upon the joint density $p\br{\rho,\bth}$:
\begin{align}
    H\br{\rho,\bth} = -\log p\br{\rho,\bth} &= -\log p\br{\rho \vert \bth} - \log p\brbth\\
    &= T\br{\rho\vert \bth} + V\brbth \text{.}
\end{align}
The term $T\br{\rho \vert \bth} = -\log p\br{\rho \vert \bth}$ is usually called kinetic energy and $V\brbth = - \log p\brbth$ is called potential energy. To generate transitions to a new state, first, a value for the momentum is drawn independently from the current $\bth$; then, Hamilton's equations are adopted to describe the evolution of the joint system $\br{\rho,\bth}$, i.e.:
\begin{align}
    \frac{d \bth}{dt} &= +\frac{\partial H}{\partial \rho} = +\frac{\partial T}{\partial \rho}\text{,}\\
    \frac{d \rho}{dt} &= -\frac{\partial H}{\partial \bth} = -\frac{\partial T}{\partial \bth} - \frac{\partial V}{\partial \bth}\text{.}
\end{align}
By having the momentum density being independent of the target density, $p\br{\rho\vert\bth} = p\br{\rho}$, $\partial T/\partial \bth = \bm{0}$, the transitions are governed by the derivatives
\begin{align}
    \frac{d \bth}{dt} &= \frac{\partial H}{\partial \rho}\text{,} \\
    \frac{d \rho}{dt} &= - \frac{\partial V}{\partial \bth}\text{.}
\end{align}
Note that $- \partial V/\partial \bth$ is simply the gradient of the negative loglikelihood, which can be computed using automatic differentiation.
The main difficulty is the simulation of the Hamiltonian dynamics, for which there is a variety of approaches \cite[see, e.g.][]{leimkuhler_reich_2005,berry2015simulating,hoffman2014no}. Yet, to solve the above system of differential equations, a leapfrog integrator is generally used due to its simplicity and volume-preservation and reversibility properties  \citep{neal2011mcmc}. The leapfrog integrator is a numerically stable integration algorithm specific to Hamiltonian systems.
It discretizes time using a step size $\eps$ and alternates half-step momentum updates and full-step parameter updates:
\begin{align}
    \rho &= \rho -\frac{\eps}{2} \frac{\partial V}{\partial \bth} \text{,}\\
    \bth &= - \bth + \eps \iS \rho\text{,}\\
    \rho &= \rho - \frac{\eps}{2} \frac{\partial V}{\partial \bth}\text{.}
\end{align}
By repeating the above steps $L$ times, a total of $L\eps$ time is simulated, and the resulting state is $\br{\rho^\star, \bth^\star}$. Note that both $L$ and $\eps$ are hyperparameters, and their tuning is often difficult in practice. In this regard, see the Generalized \acrshort{hmc} approach of \citep{horowitz1991generalized} and developments aimed at resolving the tuning of the leapfrog iterator \citep{fichtner2020auto,hoffman2022tuning},

Instead of generating a random momentum vector right away and sampling a new state $\br{\rho^\star, \bth^\star}$, to account for numerical errors in the leapfrog integrator (an analysis in this regard is found in  \citep{leimkuhler_reich_2005}), a Metropolis-hasting step is used. The probability of accepting the proposal $\br{\rho^\star, \bth^\star}$ by transitioning from $\br{\rho,\bth}$ is
\begin{equation}
  \min \br{1, e^{-H\br{\rho,\bth}+H\br{\rho^\star, \bth^\star}}}\text{.}  
\end{equation}
If the proposal $\br{\rho^\star, \bth^\star}$ is accepted, the leapfrog integrator is initialized with a new momentum draw and $\bth^\star$; otherwise, the same $\br{\rho, \bth}$ parameters are returned to start the next iteration. The \acrshort{hmc} procedure is summarized in Algorithm \ref{alg:HMC}.
Besides the difficulty of calibrating the hyperparameters $L$ and $\eps$, \acrshort{hmc} suffers from multimodality, yet the Hamiltonian boosts the local exploration for unimodal targets.

\begin{algorithm}
\caption{\acrlong{hmc}}
\label{alg:HMC}
\begin{algorithmic}
\State Initialize $\bth_0$, $\rho_0$, set $\eps$
\For{$t = 0$ to $N_\text{max}$}
    \State Simulate $\rho \sim \N\br{0,\S}$
    \State $\br{x^{(0)}, p^{(0)}} = \br{x_{i-1},p}$
\For{$j = 1$ to $L$}
    \State $\rho^{(j-\frac{1}{2}\eps)} = \rho^{(j-1)} - \frac{\eps}{2}\left. \nabla V\brbth\right\vert_{\bth = \bth_{j-1}}
    $
    \State $\bth^{(j)} = \bth^{(j-1)} + \eps \iS \rho^{(j-\frac{1}{2})}$
    \State $\rho^{(j)} = \rho^{(j-\frac{1}{2})} - \frac{\eps}{2}\left. \nabla V\brbth\right\vert_{\bth = \bth_{j}}$
\EndFor
\State $\br{x^\star, p^\star} = \br{x^{(L)},p^{(L)}}$
\State Simulate $\alpha \sim \mathcal{U}\sbr{0,1}$
\If{$\alpha < \min \br{1, \exp\br{H\br{\rho,\bth}-H\br{\rho^\star, \bth^\star}}}$}
\State $\br{x_t, p_t} = \br{x^\star, p^\star}$
\Else
\State $\br{x_t, p_t} = \br{x_{t-1}, p_{t-1}}$
\EndIf
\EndFor
\State \Return $\left\lbrace\bth_t,\rho_t\right\rbrace_{t=B}^{N\text{max}}$
\end{algorithmic}
\end{algorithm}

\section{\SectionNameWithGlossary{mcd}}
\acrfull{mcd} is an indirect method for Bayesian inference. Dropout has been earlier proposed as a regularization method for avoiding overfitting and improving neural networks' predictive performance  \citep{srivastava2014dropout}. This is achieved by applying a multiplicative Bernoulli noise on the neurons constituting the layers of the network. This corresponds to randomly switching off some neurons at each training step. The dropout rate sets the probability $p_i$ of a neuron $i$ being switched off. Though Bernoulli noise is the most common choice, note that other types of noise can be as well adopted \citep[e.g.][]{shen2018continuous}.
Neurons are randomly switched off only in the training phase, and the very same network configuration in terms of the activated and disabled neurons is used during backpropagation for computing gradients for weights' calibration. On the other hand, all the neurons are left activated for predictions.
Though it is intuitive that the above procedure implicitly connects to model averaging across different randomly pruned architectures obtained from a certain DL network, the exact connection between MC dropout and Bayesian inference follows a quite elaborated theory.

\citep{gal2016dropout} shows that a neural network of arbitrary depth and non-linearity with dropout applied before every single layer is mathematically equivalent to an approximation to the probabilistic deep \acrfull{gp} model \citep{damianou2013deep}, and  \citep{jakkala2021deep} for a recent survey. That is, the dropout objective minimizes the \acrshort{kl} divergence between a certain approximate variational model and the deep \acrshort{gp}. A treatment limited to multi-layer perceptron networks is provided in \citep{gal2015dropout}.

With $\hat{\by}$ being the output of a Neural Network with $L$ layers whose loss function is $E$, for each layer $i=1,\dots,L$ let $W_i$ denote the corresponding weight matrix of dimension $K_i \times K_{i-1}$, and $\bm{b}_i$ the bias vector of dimension $K_i$. 
Be $\by_n$ the target for the input $\bx_n$ for $n=1,\dots,N$ and denote the input and output sets respectively with $\D_x$ and $\D_y$. A typical optimization objective includes a regularization term weighted by some decay parameter $\lam$, that is
\begin{equation}\label{eq:DR:lb_dropout}
\LB_{\text{dropout}} = \frac{1}{N}\sum_{n=1}^N E\br{\by_n,\hat{\by}_n} +\lam \sum_{i=1}^L\br{\norm{W_i}^2_2+\norm{b_i}^2_2} \text{.}
\end{equation}

Now consider a deep Gaussian process for modeling distributions over functions corresponding to different network architectures. Assume its covariance is of the form 
\begin{equation}
 K\br{\bx,\by} = \int p\br{\bm{w}} p\br{b} \sigma\br{\bm{w}^\top\bx + b}  \sigma\br{\bm{w}^\top\by + b} \: \d\bm{w} \: \d b\text{,}   
\end{equation}
where $\sigma\br{\cdot}$ is an element-wise non-linearity, and $p\br{\bm{w}}$, $p\br{b}$ distributions.
Now let $W_i$ be a random matrix of size $K_i \times K_{i-1}$ for each layer $i$, be $\bm{\omega} = \cbr{W_i}_{i=1}^L$. The predictive distribution of the deep \acrshort{gp} model can be expressed as
\begin{equation}
p\br{\by_n\vert \bx_n, \D_x,\D_y} = \int p\br{\by_n\vert \bx_n,\bm{\omega}}p\br{\bm{\omega} \vert \D_x,\D_y} \: \d \bm{\omega}\text{,}
\end{equation}
where $p\br{\bm{\omega}|\vert \D_x,\D_y}$ is the posterior distribution. $p\br{\by_n\vert \bx_n,\bm{\omega}}$ is determined by the likelihood, while
$\hat{\by}_n$ is a function of $\bx_n$ and $\bm{\omega}$:
\begin{align}
p\br{\by_n\vert \bx_n,\bm{\omega}} &= \N\br{\by_n;\hat{\by}_n,I/\tau}\text{,}\\
\hat{\by}\equiv \hat{\by}\br{\bx_n,\bm{\omega}} &= \sqrt{\frac{1}{K_L}}W_L\sigma\br{\dots \sqrt{\frac{1}{K_1}}W_2\sigma\br{W_1\bx_n+\bm{m}_1}\dots}\text{.}
\end{align}
$\bm{m}_i$ are vectors of size $K_i$ for each \acrshort{gp} layer.
For the intractable posterior $p\br{\bm{\omega} \vert \D_x,\D_y}$, \citep{gal2016dropout} uses the variational approximation $q\br{\bm{\omega}}$ defined as
\begin{align}
    \bm{\omega} &= \left\lbrace W_i \right\rbrace_{i=1}^L,\label{eq:DR:variat}\\
    W_i &= M_i \,\text{diag}\br{\sbr{\bm{z}_{i,j}}_{j=1}^{K_i}},\\
    z_{i,j} &\sim \text{Bernoulli}\br{p_i}, \label{eq:DR:variat_3}\\
    &\text{\quad \, for } i=1,\dots,L \text{, } j=1,\dots,K_{i-1},\nonumber
\end{align}
where the collection of probabilities $p_i$ and matrices $M_i$, $i=1,\dots,L$ constitute the variational parameter.
Thus, $q$ stands as a distribution over (non-random) matrices whose columns are randomly set to zero, and $z_{i,j} =0$ implies that the unit $j$ in layer $i-1$ is dropped as an input to layer $i$.
For minimizing the \acrshort{kl} divergence form $q$ to $p\br{\bm{\omega} \vert \D_x,\D_y}$, the objective corresponds to
\begin{equation}\label{eq:DR:KL}
-\int q\br{\bm{\omega}} \log p\br{\D_y\vert \D_x, \bm{\omega}} + KL\br{q\br{\bm{\omega}} \vert \vert p\br{\bm{\omega}}}\text{,}
\end{equation}
By use of Monte Carlo integration and some further approximations \citep[see][for details]{gal2016dropout}, the objective reads
\begin{equation}\label{eq:DR:lb_vi}
\LB \propto \frac{1}{\tau N}\sum_{n=1}^N -\log p\br{\by_n\vert \bx_n, \hat{\bm{\omega}}_n}+ \sum_{i=1}^L \br{\frac{p_i l^2}{2 \tau N}\norm{M_i}_2^2+\frac{l^2}{2\tau N}\norm{\bm{m}_i}^2_2}
\end{equation}
which, up to the constant $\frac{1}{\tau N}$, is a feasible and unbiased \acrshort{mc} estimator of \eq{eq:DR:KL} where $\hat{\bm{\omega}}$ denotes a single \acrshort{mc} draw from the posterior $\hat{\bm{\omega}}_n \sim q \br{\bm{\omega}}$.
By taking $E\br{\by_n,\hat{\by}_n} = -\log p\br{\by_n\vert \bx_n, \hat{\bm{\omega}}_n}/\tau$  \eq{eq:DR:lb_vi} and \eq{eq:DR:lb_dropout} are equivalent for an appropriate choice of the hyperparameters $\tau$ and $l$. This shows that the minimization of the loss in \eq{eq:DR:lb_dropout} with dropout is equivalent to minimizing the \acrshort{kl} divergence from $q$ to $p\br{\bm{\omega} \vert \D_x, \D_y}$, thus performing \acrshort{vi} on the deep Gaussian process.

With an \acrshort{sgd} approach, one can maximize the above \acrshort{lb} and estimate the variational parameters from which one can simply obtain samples from the predictive distribution $q\br{\by^\star \vert \bx^\star}$, and approximate its mean by the naive \acrshort{mc} estimator:
\begin{equation}
\E_{q\br{\by^\star \vert \bx^\star}}\br{\by^\star} \approx \frac{1}{N_s} \sum_{s=1}^{N_s} \hat{\by}^\star \br{\bx^\star, \bm{\omega}^s = \left\lbrace W^s_1,\dots,W^s_L\right\rbrace}\text{.}
\end{equation}
$\bx^\star$ denotes a new observation, not in $\D_x$, for which the corresponding prediction is $\hat{\by}^\star$. I.e., the predictive mean is
obtained by performing $N_s$ forward passes through the network with Bernoulli realizations $\cbr{\bm{z}^s_1,\dots,\bm{z}^s_L}_{s=1}^{N_s}$ with $\bm{z}^s_i = [\bm{z}^s_{i,j}]_{j=1}^{K_i}$ for $s=1,\dots,N_s$, giving $\cbr{W^s_1,\dots,W^s_L}_{s=1}^{N_s}$. Such average predictions are generally referred to as \acrshort{mc} dropout estimates. Similarly, by simple moment-matching, one can estimate the predictive variance and 
higher-order statistics synthesizing the properties of $q\br{\by^\star \vert \bx^\star}$.

The predictive distribution is, in general, a multi-modal distribution resulting from superposing bi-modal distributions on each weight matrix column. This constitutes a drawback of \acrshort{mcd}, as well the implicit \acrshort{vi} on a \acrshort{gp}. Furthermore, the \acrshort{vi} approximation in eqs.\, \eqref{eq:DR:variat}-\eqref{eq:DR:variat_3} may be adequate or not. It is clear that even though \acrshort{mcd} is a possibility for \acrshort{vi} in deep-learning models, it is constrained by the very specific form in \eq{eq:DR:variat} of the variational posterior that implicitly corresponds to performing \acrshort{vi} on a deep \acrshort{gp}. Furthermore, there is evidence that \acrshort{mcd} does not fully capture uncertainty associated with model predictions \citep{chan2020unlabelled}, and there are issues related to the use of improper priors and singularity of the approximate posterior. The latter ones are addressed and explored in \citep{hron2018variational}, suggesting the use of the so-called Quasi-\acrshort{kl} divergence as a remedy. Clearly, high dropout rates drive the convergence rate slow, expand the network training time, and can cause important training data to be missed or given little relative importance.
However, compared to the traditional approach for neural networks, applying dropout places no additional effort and is often of faster training than other \acrshort{vi} methods. Furthermore, if a network has been trained with dropout, only by including an additional form of regularization acting as a prior turns the \acrshort{ann} into a \acrshort{bnn}, without requiring re-estimation \citep{jospin2022hands}.

\section{\SectionNameWithGlossary{bbb}}
A common approach for estimating the variational posterior over the networks' weights is the \acrfull{bbb} method of \cite{blundell2015weight}, perhaps a breakthrough in probabilistic deep-learning as a practical solution for Bayesian inference.

The key argument in \citep{blundell2015weight} is the use of the local reparametrization trick under which the derivative of an expectation can be expressed as the expectation of a derivative.
It introduces a random variable $\bm{\eps}$ having a probability density given by $q\br{\bm{\eps}}$ 
and a deterministic transform  $t\br{\bth,\bm{\eps}}$ such that $\bm{w} = t\br{\bth,\bm{\eps}}$. 
The main idea is that the random variable $\bm{\eps}$ is a source of noise that does not depend on the variational distribution, and the weights $\bm{w}$ are sampled indirectly as a deterministic transformation of $\bm{\eps}$, leading to a training algorithm that is analogous to that used in training regular networks.
Indeed, by writing $\bm{w}$ as $\bm{w} = t\br{\bth,\bm{\eps}}$, in place of evaluating 
\begin{equation}
\frac{\partial}{\partial \bth} \E_{q\br{\bm{w}\vert \bth}} \sbr{f\br{\bm{w},\bth}} = \frac{\partial}{\partial \bth} \int q\br{\bth\vert \bm{w}} f\br{\bm{w}, \bth} \d\bm{w} \text{,}
\end{equation}
which can be complex and rather tedious, under the assumption  $q\br{\bm{\eps}}d\bm{\eps} = q\br{\bm{w}\vert \bth}\d\bm{w}$, \cite{blundell2015weight} prove that 
\begin{align}
 \frac{\partial}{\partial \bth} \E_{q\br{\bm{w}\vert \bth}} \sbr{f\br{\bm{w},\bth}} &= \E_{q\br{\bm{\eps}}}\sbr{\frac{\partial}{\partial \bth} f\br{t\br{\bth,\bm{\eps}},\bth}}\label{eq:BBB_1} \\ 
 &= \E_{q\br{\bm{\eps}}} \sbr{\frac{\partial f\br{\bm{w},\bth}}{\partial \bm{w}} \frac{\partial \bm{w}}{\partial \bth} + \frac{\partial f\br{\bm{w},\bth} }{\partial \bth}} \text{.} \label{eq:BBB_2}
\end{align}
With $f\br{\bm{w},\bth} = \log q\br{\bm{w}\vert \bth} - \log p\br{\bm{w}}p\br{y\vert \bm{w}}$, the right side of \eq{eq:BBB_1} and \eq{eq:BBB_2} provide an alternative approach for the estimation of the gradients of the cost function with respect to the model parameters.

In fact, upon sampling $\bm{\eps}$ and obtaining $\bm{w}$, $\log q\br{\bm{w}\vert \bth} - \log p\br{\bm{w}}p\br{y\vert \bm{w}}$ is a stochastic approximation of the \acrshort{vi} objective $\KL\sbr{q\br{\bm{w}\vert \bth} \vert \vert p\br{\bm{w}\vert \D}} =  \E_{q\br{\bm{w}\vert \bth}} \sbr{\log q\br{\bm{w}\vert \bth} - \log p\br{\bm{w}}p\br{\D\vert \bm{w}}} $ to be minimized.

The sampled value $\bm{\eps} \sim q\br{\bm{\eps}}$, resampled at each iteration, is independent of the variational parameters, while $\bm{w}$ is not directly sampled but here it is a deterministic function of $\bm{\eps}$. Given $\bm{\eps}$, all the quantities in the square bracket of \eq{eq:BBB_2} are non-stochastic, enabling the use of backpropagation. A single draw for $\bm{\eps}$ approximates the right side of \eq{eq:BBB_1}, and suffices for providing an unbiased stochastic gradient estimation of the relevant gradient on the left side. 
Equation \eqref{eq:BBB_1} makes explicit the possibility of using automatic differentiation to compute the gradient of $f$ with respect to the parameter $\bth$. By using a single sampled draw $\bth$ for approximating the expectation on the right side of \eq{eq:BBB_1}, the only parameter in the loss is $\bth$, and the use of backpropagation for evaluating the gradients is straightforward. 
Equation \eqref{eq:BBB_2} instead employs backpropagation in the \enquote{usual} sense, involving gradients of the cost with respect to the network parameters $\bm{w}$, further rescaled by $\partial \bm{w}/\partial \bth$ and shifted by $\partial f\br{\bm{w},\bth}/\partial \bth$. Equation \eqref{eq:BBB_2} concerns the usual backpropagation computations in terms of the network's weights, the specific form of the partial derivative with respect to $\bth$ that the choice of $t$ implies, while the last term depends on the chosen form of the variational posterior only ($\bm{w}$ is here not seen as a function of $\bth$, as the form of \eq{eq:BBB_2} results from applying the multi-variable chain rule). 
This results in a general framework for learning the posterior distribution over the network's weights. The following algorithm \ref{alg:BBB_general} summarizes the \acrshort{bbb} approach.

\begin{algorithm}
\caption{\acrlong{bbb}}
\label{alg:BBB_general}
\begin{algorithmic}
\Repeat
    \State Simulate $\bm{\eps} \sim q\br{\bm{\eps}}$
    \State $\bm{w} = t\br{\bth,\bm{\eps}}$
    \State $f\br{\bm{w},\bth} = \log q\br{\bm{w}\vert \bth} - \log p\br{\bm{w}}p\br{\D \vert \bm{w}}$
    \State $\nabla_\bth f = \text{backprop}_\bth \br{f}$
    \State $\bth = \bth -\beta \nabla_\bth f$
\Until{stopping criterion is met}
\end{algorithmic}
\end{algorithm}

Algorithm \ref{alg:BBB_general} is initialized by preliminary setting the initial values of the variational parameter $\bth$ and, of course, by specifying the form of the prior and the posterior along with the form of the likelihood involving the outputs of the forward pass obtained from the specified underlying network structure.
The update is very similar to the one employed in standard non-Bayesian settings, where standard optimizers such as \acrshort{adam} are applicable. It is the applicability of standard optimization algorithms and the use of classic backpropagation that constitute the major breakthrough element in \acrshort{bbb}, making it a feasible approach for Bayesian learning.

To make the description more explicit and aligned with the following sections, we present the case where the variational posterior is a diagonal Gaussian with mean $\bmu$ and covariance matrix $\sigma^2 I$. In this case, the transform $t$ takes the simple and convenient form  
\begin{equation}
\bm{w} = t\br{\bth,\bm{w}} = \bmu + \sigma \bm{\eps}\text{.}
\end{equation}
As $\sigma$ is required to be always non-negative, \citep{blundell2015weight} adopts the reparametrization $\sigma = \log\br{1+\exp\br{\rho}}$ and the variational posterior parameter $\bth = \br{\bmu,\rho}$. In this case, Algorithm \ref{alg:BBB_gaussian} summarizes the \acrshort{bbb} approach.

\begin{algorithm}
\caption{\acrlong{bbb} for a diagonal Gaussian variational posterior}
\label{alg:BBB_gaussian}
\begin{algorithmic}
\Repeat
    \State Simulate $\bm{\eps} \sim \N\br{\bm{0},I}$
    \State $\bm{w} = \bmu + \log\br{1+\exp\br{\rho}}\odot \bm{\eps}$
    \State $f\br{\bm{w},\bth} = \log q\br{\bm{w}\vert \bth} - \log p\br{\bm{w}}p\br{\D\vert \bm{w}}$
    \State $\nabla_\bmu f = \frac{\partial f\br{\bm{w},\bth}}{\partial \bm{w}} + \frac{\partial f\br{\bm{w},\bth}}{\partial \bmu}$
    \State $\nabla_\rho f = \frac{\partial f\br{\bm{w},\bth}}{\partial \bm{w}} \frac{\bm{\eps}}{1+\exp\br{-\rho}}+ \frac{\partial f\br{\bm{w},\bth}}{\partial \rho}$
    \State $\bmu = \bmu -\beta \nabla_\mu f$
    \State $\rho = \rho -\beta \nabla_\mu f$
\Until{stopping criterion is met}
\end{algorithmic}
\end{algorithm}

As for Algorithm \ref{alg:BBB_gaussian}, one may backpropagate the gradients of $f$ w.r.t. $\bmu$ and $\rho$ directly. Alternatively, as for Algorithm \ref{alg:BBB_general}, one may use backpropagation for computing the gradients $\partial f\br{\bm{w},\bth}/\partial \bm{w}$, which are furthermore shared across the updates for $\bmu$ and $\rho$, or, if preferred, adopt a general automatic differentiation setup, if, e.g., the form of the variational likelihood does not allow for a simple analytic form of the gradient.


\section{Exponential family and Natural gradients}\label{sec:ExpFamily}

Assume $q_\bl\brbth$ belongs to an exponential family distribution. Its probability density function is parametrized as
\begin{equation}\label{eq:exp_fam_representation}
    q_\bl\brbth = h\brbth \exp\br{\phi\brbth^\top \bl - A\br{\bl}}\text{,}
\end{equation}
where $\bl \in \Omega$ is the natural parameter, $\phi\brbth$ the sufficient statistic. $A\br{\bl} = \log \int h\brbth \exp(\phi\brbth^\top \bl) d\nu$ is the log-partition function, determined upon the measure $\nu$, $\phi$ and the function $h$. The natural parameter space is defined as $\Omega = \{  \bl \in \mathbb{R}^d: A\br{\bl} < +\infty \}$.
When $\Omega$ is a non-empty open set, the exponential family is referred to as regular. Furthermore, if there are no linear constraints among the components of $\bl$ and $\phi\brbth$, the exponential family in \eq{eq:exp_fam_representation} is said of minimal representation. Non-minimal families can always be reduced to minimal families through a suitable transformation and reparametrization, leading to a unique parameter vector $\bl$ associated with each distribution \citep{wainwright2008graphical}.
The mean (or expectation) parameter $\bm{m} \in \mathcal{M}$ is defined as a function of $\bl$, $\bm{m}\br{\bl} = \Eq{q_\bl}\sbr{\phi\brbth} = \nabla_\bl A\br{\bl}$. 
Moreover, for the Fisher Information Matrix $\I_\bl = -\Eq{q_\bl}\sbr{\nabla_\bl^2 \log q_\bl\brbth}$ it holds that $\I_\bl =\nabla^2_\bl A\br{\bl} = \nabla_\bl \bm{m} \text{.}$
Under minimal representation, $A\br{\bl}$ is convex, thus the mapping $\nabla_\bl A = \bm{m}:\Omega \rightarrow \mathcal{M}$ is one-to-one, and $\I_\bl$ is positive definite and invertible \citep{nielsen2009statistical}. $\mathcal{M}$ denotes the set of realizable mean parameters.
Therefore, under minimal representation we can express $\bl$ in terms of $\bm{m}$ and thus $\LB\br{\bl}$ in terms of $\LB\br{\bm{m}}$ and vice versa \citep{khan2018fastyetsimple}. 

\begin{example}[The Gaussian distribution as an exponential-family member]
The multivariate Gaussian distribution $\mathcal{N}\br{\bmu,\S}$ with $k$-dimensional mean vector $\bmu$ and covariance matrix $\S$ can be seen as a member of the exponential family (eq. \eqref{eq:exp_fam_representation}). Its density reads
\begin{equation}
    q_\bl\brbth = \br{2\pi}^{k/2}\exp\{\phi\brbth^\top \bl - \frac{1}{2}\bmu^\top \iS \bmu -\frac{1}{2}\log \vert \S\vert\} \text{,}
\end{equation}
where
\begin{equation}
    \phi\brbth =\begin{bmatrix}
           \theta \\
           \theta\theta^\top
         \end{bmatrix} \text{,}\quad
    \bl =\begin{bmatrix}
           \bl_1 \\
           \bl_2
         \end{bmatrix} =\begin{bmatrix}
           \iS \bmu \\
           -\frac{1}{2}\iS
         \end{bmatrix} \text{,}\quad
    \bm{m} = \begin{bmatrix}
           \bm{m}_1 \\
           \bm{m}_2
         \end{bmatrix} = \begin{bmatrix}
          \bmu \\
          \S + \bmu\bmu^\top
         \end{bmatrix} \text{,}                  
\end{equation}
and $A\br{\bl} = -\frac{1}{4}\bl_1^\top\bl_2^{-1} \bl_1-\frac{1}{2} \log \br{-2\bl_2} $. On the other hand, $\bz = \sbr{\bz_1^\top, \bz_2^\top}^\top$ with $\bz_1 = \bmu = \bm{m}_1$ and $\bz_2 = \S = \bm{m}_2 - \bmu \bmu^\top$, constitutes the common parametrization of the multivariate Gaussian distribution in terms of its mean and variance-covariance matrix. 
\end{example}

By applying the chain rule, $\nabla_\bl\LB = \nabla_\bl \bm{m} \nabla_{\bm{m}} \LB = \nabla_\bl \br{\nabla_\bl A}\LB = \nabla^2_\bl A\br{\bl} \LB =  \mathcal{I}_\bl \nabla_{\bm{m}} \LB$, from which
\begin{equation} 
 \natgrad_\bl \LB = \I^{-1}_\bl \nabla_\bl\LB = \nabla_{\bm{m}} \LB\text{,} \label{eq:natgrad_lambda_grad_m}
\end{equation}

The quantity $\natgrad_\bl \LB$ is referred to as the natural gradient of $\LB$ with respect to $\lam$ and it is obtained by pre-multiplying the Euclidean gradient by the inverse of the \acrshort{fim} (parametrized in terms of $\bl$). In general, $\LB$ can be a generic function whose derivative with respect to a parameter $\bl$ (not necessarily the natural parameter) exists. The standard reference for natural gradients computation is the seminal work of \cite{amari1998natural}. Within a \acrshort{sgd} context, the application of simple Euclidean gradients is problematic as it ignores the information geometry of the distribution $q_\bl$. Euclidean gradients implicitly rely on the Euclidean norm to capture the dissimilarity between two distributions which can be a quite poor dissimilarity measure \citep{khan2018fastyetsimple}.
In fact, the \acrshort{sgd} update can be obtained by writing
\begin{equation}
    \bl_{t+1} = \argmin_{\bl} \bl^\top \sbr{\nabla_\bl \LB\br{\bl_t}} -\frac{1}{2\beta} \vert\vert \bl-\bl_t \vert \vert^2
\end{equation}
and setting to zero its derivative. Although the above implies that $\bl$ moves in the direction of the gradient, it remains close to the previous $\bl_t$ in terms of Euclidean distance. As $\bl$ is a parameter of a distribution, the adoption of the Euclidean measure is misleading. An Exponential family distribution induces a Riemannian manifold with a metric defined by the FIM \citep{khan2018fastyetsimple}. By replacing the Euclidean metric with the Riemannian one,
\begin{equation}
    \bl_{t+1} = \argmin_{\bl} \bl^\top \sbr{\nabla_\bl \LB\br{\bl_t}} -\frac{1}{2\beta} \br{\bl-\bl_t}^\top \I_\bl \br{\bl-\bl_t}
\end{equation}
the resulting update is indeed expressed in terms of the natural parameter:
\begin{equation}
    \bl_{t+1} = \bl_t + \beta \I^{-1}_\bl  \nabla_\bl \LB\br{\bl_t}\text{,}
\end{equation}
generally referred to as natural gradient update. More in general, one could replace the Euclidean distance with a proximity function such as the Bregman divergence and obtain richer classes of \acrshort{sgd}-like updates, like mirror descent (which can be interpreted as natural gradient descent), see, e.g., \citep[][]{nielsen2020note}.
A very interesting point on the limitations of plain gradient search is made in \citep{wierstra2014natural} concerning the impossibility of locating, even in a one-dimensional case, a quadratic optimum. The example provided therein involves the Gaussian distribution, pivotal in \acrshort{vi}. For an one-dimensional Gaussian distribution with mean $\mu$ and standard deviation $\sigma$, the gradient of $\LB$ with respect to the parameters $\mu$ and $\sigma$ lead to the following SGD updates:
\begin{align}
  \mu &= \mu +\beta\nabla_\mu\LB = \mu + \beta\frac{z-\mu}{\sigma^2}\\
  \sigma &= \sigma + \beta\nabla_\sigma \LB = \frac{\br{z-\mu}^2 - \sigma^2}{\sigma^3}\text{.}
\end{align}

For the updates to converge and the optimum to be precisely located, $\sigma$ must decrease (i.e., the distribution shrinks around $\mu$). The fact that $\sigma$ appears in the denominator of both the updates is problematic: as it decreases, the variance of the updates increases as $\Delta_\mu \propto \frac{1}{\sigma}$ and $\Delta_\sigma \propto \frac{1}{\sigma}$. The updates become increasingly unstable, and a large overshooting update makes the search start all over again rather than converging. Increased population size and small learning rates cannot avoid the problem. The choice of the starting value is problematic, too: starting with $\sigma >>1$ makes the updates minuscule; conversely, $\sigma <<1$ makes them huge and unstable. \citep{wierstra2014natural} discusses how the use of natural gradients fixes this issue that, e.g., may arise with \acrshort{bbvi}.

Algorithm \ref{alg:wierstra} summarizes the generic scheme upon the implementation of a natural gradient update. In Algorithm \ref{alg:wierstra}, $\bz$ denotes a generic variational parameter, e.g., the natural parameter or not, while methods for evaluating $\nabla_\th \LB$, $\I$, and efficiently computing its inverse $\iI$ are discussed in the following sections.

\begin{algorithm}\label{alg:wierstra}
\caption{\acrlong{sgd} with natural gradients}
\begin{algorithmic}
\Repeat
    \State Compute: $\nabla_\bz \LB$, $\I$
    \State $\bz = \bz +\beta \I^{-1} \nabla_\bz \LB$
\Until{stopping condition is met}
\end{algorithmic}
\end{algorithm}


\section{Black-Box methods}
A major issue in \acrshort{vi} is that it heavily relies upon model-specific computations, on which a generalized, ready-to-use, and plug-and-play optimizer is difficult to design. Black-box methods aim at providing solutions that can be immediately applied to a wide class of models with little effort. In the first instance, the ubiquitous use of model's gradients that traditional \acrshort{ml} and \acrshort{vi} approaches rely upon struggles with this principle. As \citep{ranganath2014black} describes, for a specific class of models, where conditional distributions have a convenient form and a suitable variational family exists, \acrshort{vi} optimization can be carried out using closed-form coordinate ascent methods \citep{ghahramani2000propagation}. In general, there is no close-form solution resulting in model-specific algorithms \citep{jaakkola1997variational,blei2007correlated,braun2010variational} or generic algorithms that involve model-specific computations \citep{knowles2011non, paisley2012variational}. As a consequence model assumptions and model-specific functional forms play a central role, making \acrshort{vi} practical.
The general idea of Black-box VI is that of rewriting the gradient of the \acrshort{lb} objective as the expectation of an easy-to-compute function of the latent and observed variables. The expectation is taken with respect to the variational distribution, and the gradient is estimated by using stochastic samples from it in a \acrshort{mc} fashion. Such stochastic gradients are used to update the variational parameters following an \acrshort{sgd} optimization approach. Within this framework, the end-user is required to develop functions only for evaluating the model log-likelihood, while the remaining calculations are easily implemented in libraries of general use applicable to several classes of models.
Black-box VI falls within stochastic optimization where the optimization objective is the maximization of the LB using noisy, unbiased, estimates of its gradient. As such, variance reduction methods have a major impact on stability and convergence, among them control variates are the most effective and of immediate implementation.

\subsection{\SectionNameWithGlossary{bbvi}}
\acrfull{bbvi} optimizes the \acrshort{lb} with stochastic optimization, through an unbiased estimator of its gradients obtained from samples from the variational posterior \citep{ranganath2014black}. By using the \acrshort{lb} definition and the log-derivative trick on the gradient of the LB with respect to the variational parameter, $\nabla_\bz \LB$ can be expressed as
\begin{equation}
\nabla_\bz \LB = \E_q\sbr{\nabla_\bz \log q\br{\bth \vert \bz}\br{\log p\br{\D,\bth} - \log q\br{\bth\vert \bz}}}\text{,}  
\end{equation}
where $\bz$ denotes the parameter of the variational distribution $q_\bz$.
The above expression rewrites the gradient as an expectation of a quantity that does not involve the model's gradients but only those of $\log q\br{w \vert \bz}$. A naive noisy unbiased estimate of the gradient of the LB is immediate to obtain with $N_s$ samples obtained from the variational distribution,
\begin{equation}
\nabla_\bz \LB = \frac{1}{N_s}\sum_{s=1}^{N_s} \nabla_\bz \log q\br{\bth_s\vert \bz}\sbr{\log p\br{\D,\bth_s} - \log q\br{\bth_s\vert \bz}} \text{,}
\end{equation}
where $\bth_s \sim q\br{\bth\vert \bz}$. The above \acrshort{mc} estimator enables the immediate and feasible computation of the \acrshort{lb} gradients as, given a sample $\bth_s$, $\log q\br{\bth_s\vert \bth}$ is a quantity that solely depends on the form of the variational posterior and can be of simple form. On the other hand, $\log p\br{\D,\bth} - \log q\br{\bth\vert \bth}$ is immediate to compute as it only requires evaluating the logarithm of the joint $p\br{\D,\bth_s}$ and the density of the variational distribution in $\bth_s$. This process is summarized in Algorithm \ref{alg:BBVI_1}. If sensible, one may assume that $\log p\br{\D,\bth} = \log p\br{\D\vert \bth} p\br{\bth}$ but this is not explicitly required as of \citep{ranganath2014black}: there are no assumptions on the form of the model; the approach only requires the gradient of the variational likelihood with respect to the variational parameters to be feasible to compute.

\begin{algorithm}
\caption{\acrlong{bbvi}}
\label{alg:BBVI_1}
\begin{algorithmic}
\Repeat
    \State Simulate $\bth_s \sim q_\bz$, for $s=1,\dots,N_s$
    \State $\nabla_\bz \LB = \frac{1}{N_s}\sum_{s=1}^{N_s} \nabla_\bz \log q\br{\bth_s\vert \bz}\sbr{\log p\br{\D\vert \bth_s}p\br{\bth_s} - \log q\br{\bth_s\vert \bz}}$
    \State $\bz = \bz +\beta \nabla_\bz \LB$
\Until{stopping criterion is met}
\end{algorithmic}
\end{algorithm}

In \citep{ranganath2014black}, the authors employ an adaptive learning rate satisfying the Robbins Monroe conditions $\sum_t \beta_t = \infty$ and $\sum_t \beta^2_t < \infty$, and for controlling the variance of the stochastic gradient estimator adopt Rao-blackewllization \citep{rao1945information,blackwell1947conditional,robert2021rao} and use the of control variates \citep[e.g.][Ch.\, 3]{lemieux2014control,robert1999monte} within Algorithm \ref{alg:BBVI_1}.

\subsection{\SectionNameWithGlossary{ngbbvi}}
We shall review the approach of \cite{trusheim2018boosting} boosting \acrshort{bbvi} with natural gradients, referred to as \acrfull{ngbbvi}.
The \acrshort{fim} corresponds to the outer product of the score function with itself (see Section \ref{sec:ExpFamily}) and is furthermore equal to the second derivative of the \acrshort{kl} divergence to the approximate posterior $q\br{x\vert \bz}$:
\begin{align}\label{eq:ngbbvi:fim}
    F\br{\bz} &= \left.\frac{d^2 \KL\sbr{q_\bz\brbth \vert \vert q_{\hat{\bz}}\brbth }}{\br{d\bz}^2}\right\vert_{\hat{\bz}=\bz} = \Eq{q_\bz}\sbr{\nabla_\bz \log q_\bz\brbth \nabla_\bz \log q_\bz\brbth^\top}\text{.}
\end{align}
For the practical implementation, \citep{trusheim2018boosting} uses a mean-field restriction on the variational model, i.e. the joint is factorized into the product of $K$ independent terms, where each term is in general a multivariate distribution:
\begin{equation}
q_\bz\brbth = \prod_{k=1}^K q_{\bz_k}\br{\bth_k}\text{.}
\end{equation}
The above restriction is also suggested by \cite{ranganath2014black} in order to allow for Rao-Blackwellization \citep{robert2021rao} as a tool to be used in conjunction with control variates \citep[e.g.][Ch.\, 3]{lemieux2014control} for reducing the variance of the stochastic gradient estimator. Under the above assumption, the FIM simplifies to:
\begin{equation}
\I_\bz = 
\begin{cases}
\E_{q_i\br{\bth\vert \bz}}\sbr{\nabla_{\bz_i} \log q_{\bz_i}\br{\bth_i} \nabla_{\bz_i} \log q_{\bz_i}\br{\bth_i} \nabla_{\bz_i}^\top} \text{,} & i=j,\\
0 \text{,} & i\neq j,
\end{cases}
\end{equation}
which significantly simplifies the general form $q_\bz\brbth$ while implicitly enabling Rao-Blackwellization with the variable-wise local expectations and thus reducing the variance of the \acrshort{fim}, estimated via a Monte-Carlo approach. In fact, besides a few variational models it is difficult to compute the above expectations analytically so \cite{trusheim2018boosting} adopts the following naive \acrshort{mc} estimator:
\begin{equation}
\hat{\I}_\bz = 
\begin{cases}
\frac{1}{N_s}\sum_{s=1}^{N_s}\sbr{\nabla_{\bz_i} \log q_{\bz_i}\br{\bth_i^{\br{s}}} \nabla_{\bz_i} \log q_{\bz_i}\br{\bth_i^{\br{s}}}^\top}\text{,} &  i=j\\
0 \text{,} & i\neq j
\end{cases}
\end{equation}
with $\bth_i^{\br{s}} \sim q_{\bz_i} \br{\bth_i}$ denoting a sample from the $i$-th factor of the posterior mean-field approximation. Note that the above does not introduce additional computations as the score of the samples $\bth_i^{\br{s}}$ is anyway required in the computation of the LB gradient.
Furthermore, instead of using a plain \acrshort{sgd}-like update, \citep{trusheim2018boosting} adopts an \acrshort{adam}-like version, boosted with natural gradient computations. Algorithm \ref{alg:NG-BBVI} summarizes the \acrshort{ngbbvi} approach.

\begin{algorithm}[!ht]
\caption{\acrlong{ngbbvi}}
\label{alg:NG-BBVI}
\begin{algorithmic}
\Repeat
    \For{$s=1$ to $N_s$}
        \State Simulate $\bth\sbr{s} \sim q_\bz$
    \EndFor
    \State $X \leftarrow \text{subset of } \bth$, \, $Y \leftarrow \bth \setminus X$, \, $N_X = \text{numel}\br{X}$,
    \, $N_Y = \text{numel}\br{Y}$
    \For{$k=1$ to $K$}
    \State $a^\star=0$
        \For{$s=1$ to $N_X$}
            \State $h_X^{\br{s}} = \nabla_{\bz_k} \log q_{\bz_k}(X_k^{\br{s}})$
            \State $f_X^{\br{s}} = h_X^{\br{s}} \br{\nabla_{\bz_k} \log p_k(X_k^{\br{s}},\D)-\log q_{\bz_k}(X_k^{\br{s}})-a^\star}$
        \EndFor
        \State $a^\star = \text{Cov}\br{h_X,f_X}/\text{Var}\br{h_X}$
        \For{$s=1$ to $N_Y$}
            \State $h_Y^{\br{s}} = \nabla_{\bz_k} \log q_{\bz_k}(Y_k^{\br{s}})$
            \State $f_Y^{\br{s}} = h_Y^{\br{s}} \br{\nabla_{\bz_k} \log p_k(Y_k^{\br{s}},\D)-\log q_{\bz_k}(Y_k^{\br{s}}) - a^\star}$
        \EndFor
        \State $\hat{\I}_{\bz_k} = \frac{1}{N_Y} \sum_{s=1}^{N_Y} h_Y^{\br{s}} {h_Y^{\br{s}\top}}$
        \State $\hat{\nabla}_{\bz_k} \LB \leftarrow \hat{\I}^{-1}_{\bz_k} \br{\frac{1}{N_Y}\sum_{s=1}^{N_Y} f_Y^{\br{s}}}$
    \EndFor
    \State $\bm{m} = \beta_1 \bm{m} + \br{1-\beta_2} \hat{\nabla}_{\bz} \LB $
    \State $\bm{v} = \beta_2 \bm{v} +\br{1-\beta_1}\norm{\hat{\nabla}_{\bz} \LB}^2$
    \State $\hat{\bm{m}} = \frac{\bm{m}}{1-\beta_2^t}$, \, $\hat{\bm{v}} = \frac{\bm{v}}{1-\beta_2^t}$
    \State $\bz = \bz +\beta \frac{\hat{\bm{m}}}{\sqrt{\hat{\bm{v}}}+\eps}$
 
\Until{$\norm{d\bz} < \text{threshold}$}
\end{algorithmic}
\end{algorithm}

The \acrshort{ngbbvi} implementation is slightly more complex than the original \acrshort{bbvi}, see Algorithm \ref{alg:NG-BBVI}. 
The \acrshort{mc} computation involves both the black-box stochastic gradient estimation and the estimation of the optimal control variate coefficient $a^\star$. Thus the posterior samples are split into two subsets. The first one $X$ aimed at estimating $a^\star$, and the second one $Y$ at implementing the \acrshort{mc} estimators, independently from $X$, and with the control variate correction term $a^\star$ earlier computed. The computation of the \acrshort{fim} follows immediately from \eq{eq:ngbbvi:fim}, and the computation of $\nabla_{\bz_k} \LB$ is analogous to \acrshort{bbvi}. The last four lines of Algorithm \ref{alg:NG-BBVI} correspond to the implementation of the \acrshort{adam} update, operators are intended to be applied element-wise, $\beta_1$, $\beta_2$ (exponential decay rates) are typical \acrshort{adam} hyperparameters, $\eps>0$ is a small offset preventing divisions by zero.

\citep{trusheim2018boosting} differs from \acrshort{bbvi} by the use of natural gradients (and the adoption of the \acrshort{adam}-like update, though applicable to \acrshort{bbvi} as well). On the other hand, the use of control variates and Rao-Blackwellization for variance reduction is found in both \acrshort{bbvi} and \acrshort{ngbbvi}. As the natural gradient approach is preferable for the reasons discussed in Section \ref{sec:ExpFamily}, \acrshort{ngbbvi} is favored over \acrshort{bbvi}.

The use of the black-box framework for computing the gradients of the \acrshort{lb} along the \acrshort{mc} estimator for the \acrshort{fim} renders \acrshort{ngbbvi} of general applicability and not constrained to a certain form of the variational posterior. Yet the \acrshort{mc}-computations of the \acrshort{fim} are implicitly approximate, whereas for certain distributions the \acrshort{fim} computation can be carried out analytically and in an exact form. \acrshort{ngbbvi} furthermore requires the inversion of the \acrshort{fim}, which is a computational bottleneck. The following \acrshort{von} \cite{khan2018fastyetsimple}, \acrshort{vadam} \citep{khan2018fastscalable} and \acrshort{vogn} \citep{khan2018fastscalable,osawa2019practical} methods indeed fix this issue: assuming a variational posterior within the exponential distribution family, natural gradients are enabled without the direct computation of the \acrshort{fim} and its inverse.

\section{Natural gradient methods for exponential-family variational distributions}\label{sec:NB_methods}

In the following subsections, we review methods based on Natural gradients and Exponential-family variational approximations. The following techniques are built on natural parameter updates in the natural parameter space and rely on simplified but exact \acrshort{fim} computations based on the natural/expectation parameter duality (\eq{eq:natgrad_lambda_grad_m}).

\subsection{Exact gradient computations for the exponential family}
The computation of the \acrshort{fim} required in the natural gradient computation is, in general, not trivial. In a generic perspective, not bound to a specific variational form, the sampling approach for the \acrshort{fim} estimation of \citep{trusheim2018boosting} is feasible. Yet for certain distributions, namely for those in the Exponential family class, natural gradients can be computed in an exact form with an analytical solution which furthermore does not involve the computation of the \acrshort{fim}. 

The theoretic foundation of such a viable approach is provided in \citep{khan2018fastyetsimple} and traces back to \eq{eq:natgrad_lambda_grad_m}. For an Exponential family of minimal representation, the natural gradient with respect to the natural parameter $\bl$ is equal to the gradient with respect to the expectation parameter $\bm{m}$. This is a powerful result that allows the computation of the natural gradient as an Euclidean gradient, avoiding the computation of the \acrshort{fim} and its inversion. 

This section presents some baseline methods using the above duality for the natural gradient computation. Differently from \acrshort{bbb}, \acrshort{bbvi}, and \acrshort{ngbbvi} the following approaches explicitly deal with variational distributions members of the Exponential family with a focus on updating their natural parameter:
\begin{align}
\bl_{t+1} &= \bl_t +\beta \natgrad_\bl \LB\br{\bl_t} \label{eq:NATGRAD:update_lam} \\
&= \bl_t + \beta \I^{-1}_{\bl_t} \nabla_{\bl_t} \LB\br{\bl_t} = \bl_t +\beta \nabla_{\bm{m}} \LB\br{\bm{m}_t}. \label{eq:NATGRAD:update_lam_grad_m}
\end{align}
From the above updates on natural parameters, update rules for  alternative and perhaps a more usual parametrization can often be obtained, see e.g. Section \ref{sec:ngvi}.

\subsection{\SectionNameWithGlossary{ngvi}}\label{sec:ngvi}
\acrfull{ngvi} constitutes a baseline methodology for natural gradient computation under a Gaussian variational distribution, upon which several other approaches have been developed. 

The natural gradients in the natural parameter space can be computed under the expectation parametrization as Euclidean gradients. \citep{khan2018fastyetsimple} shows that such gradients are of simple form and correspond to
\begin{align}
    \natgrad_{\bl_1} \LB = \nabla_{\bm{m}_1}\LB &= \nabla_\bmu \LB -2\sbr{\nabla_\S \LB}\bmu,\label{eq:NGVI:nat_grad_lam1_m1}\\
    \natgrad_{\bl_2} \LB = \nabla_{\bm{m}_2}\LB &= \nabla_\S \LB. \label{eq:NGVI:nat_grad_lam2_m2}
\end{align}
By using the definition of natural gradients in terms of $\bmu$ and $\S$, the update in \eq{eq:NATGRAD:update_lam_grad_m} for the natural parameter $\bl_1 = \bmu$, $\bl_2 = -\frac{1}{2}\iS$ rewrites as
\begin{align}
    \iS_{t+1} &= \iS_t  \nabla_\bmu \LB -2\beta \sbr{\nabla_\S \LB}\bmu\\
    \bmu_{t+1} &= \S_{t+1}\sbr{\iS_t \bmu + \beta \br{\nabla_\bmu \LB -2\sbr{\nabla_\S \LB}\bmu} }\\
    &= \S_{t+1}\sbr{\br{\iS_t -2\beta \sbr{\nabla_\S \LB}\bmu} + \beta \nabla_\bmu \LB}\\
    & = \S_{t+1} \sbr{\iS_{t+1}\bmu_t +\beta \nabla_\bmu \LB}\\
    & = \bmu_t + \beta \S_{t+1}\sbr{\nabla_\bmu \LB}.
\end{align}
The above two constitute the \acrshort{ngvi} update rules for updating the mean $\bmu$ and covariance matrix $\S$ of the variational posterior with a natural gradient update, that however does not involve the computation of the \acrshort{fim} as it relies on Euclidean gradients.

For a diagonal covariance matrix $\S = \diag{\sigma^2}$, the corresponding \acrshort{ngvi} updates read
\begin{align}
    \sigma^{-2}_{t+1} &= \sigma^{-2}_t -2\beta \sbr{\nabla_{\sigma^2}\LB_t},\\
    \bmu_{t+1} &= \bmu_t + \beta \sigma^2_{t+1}\odot \sbr{\nabla_\bmu \LB_t}.
\end{align}

With respect to the \acrshort{ngvi} update two points are important to stress out. First, at each iteration, the update for $\bmu$ implicitly requires $\S_{t+1}$. This means that the update for $\bmu$ follows that for $\iS$ and that $\bmu$ readily uses the one-step ahead updated information on $\S$. Though it may appear counter-intuitive, \citep{lyu_black-box_2021,magris2022exact} show that this update is not optimal (in the terms therein discussed), while an update of the form $\bmu_{t+1} = \bmu_t + \beta \S_{t}\sbr{\nabla_\bmu \LB}$ would be. Also, note that the update for $\bmu$ involves $\S_{t+1}$ and not $\iS_{t+1}$, meaning that in the \acrshort{ngvi} an online inversion of $\iS$ is implicitly required at each iteration. Clearly, for the diagonal case, this is trivial and effortless to obtain.
Second, in the full-covariance case, there is no guarantee that the updates guarantee $\S$ to be a positive-definite covariance matrix. This issue is tackled in Section \ref{subsec:manifold_intro}. For the diagonal case, the constraint on $\S$ results in guaranteeing the positivity of the entries in the diagonal. This can be achieved via a proper reparametrization, e.g. \acrshort{bbvi} updates $\rho$ where $\sigma = \log\br{1+\exp\br{\rho}}$, or \citep[e.g.][]{tan2021analytic} updates the Cholesky factor. Alternatively, the learning rate can be adapted to guarantee that the step size does not drive the updates $\sigma^{-2}$ negative \citep[e.g.,][]{khan2018fastyetsimple,magris2022quasi}.

\subsection{\SectionNameWithGlossary{von}}
A computational burden in \acrshort{ngvi} is that the gradients of the \acrshort{lb}  are still required: \acrfull{von} develops on \acrshort{ngvi} but does not require the gradients of the variational objective. Furthermore, it only involves the gradient and Hessian of the model log-likelihood which can be computed with usual backpropagation.

\cite{khan2018vadam} express the lower bound as
\begin{equation}
\LB = \Eq{q_\bl}\sbr{-Nf\brbth + \log p\brbth - \log q \brbth} \text{,}
\end{equation}
where $N$ is the sample size, and $f\brbth = -\frac{1}{N} \sum_{i=1}^N\log p\br{\D_i \vert \bth}$ is negative log-likelihood of the model, i.e. standard MLE objective, where $\D_i$ denotes a data example, i.e. $\D_i = \br{\by_i,\bx_i}$. \acrshort{von} uses the theoretical results of \citep{opper2009variational,rezende2014stochastic} to express the gradients of the \acrshort{lb}  objective in terms of gradient and Hessian of $f\brbth$. By linearity of the expectation, the gradients of the $\LB$ consist of the sum of the gradients of three expectation terms, in particular:
\begin{align}
    \nabla_\bmu \E_q\sbr{f\brbth} &= \E_q\sbr{\nabla_\bth f\brbth} = \E_q\sbr{\bm{g}\brbth},\\
    \nabla_\S \E_1 \sbr{f\brbth} &= \frac{1}{2}\E_q \sbr{\nabla^2_{\bth\bth} f \brbth} = \frac{1}{2}\E_q \sbr{H\brbth},
\end{align}
where $\bm{g} = \nabla_\bth f\brbth$ and $H\brbth = \nabla^2_{\bth\bth} f \brbth$ denote the gradient and Hessian of the MLE objective, respectively. With these relations the gradients of the \acrshort{lb} objective write
\begin{align}
    \nabla_\bmu \LB &= \nabla_\bmu \E_q\sbr{-Nf\brbth + \log p\brbth - \log q \brbth}\\
    &= -\br{\E_q \sbr{N \nabla_\bth f\brbth} + 0 +\bl \bmu}\\
    &= -\br{\E_q \sbr{N \bm{g} \brbth} + \bl \bmu}
\end{align}
and
\begin{align*}
    \nabla_\S \LB &= \frac{1}{2}\E_q\sbr{-N \nabla^2_{\bth\bth} f \brbth} +0 -\frac{1}{2}\bl I +\frac{1}{2}\iS\\
    &= \frac{1}{2}\E_q\sbr{-N H\brbth} -\frac{1}{2}\bl I +\frac{1}{2}\iS.
\end{align*}
By using these gradients in the \acrshort{ngvi} update, one obtains
\begin{align}
    \bmu_{t+1} &= \bmu_t -\beta \S_{t+1}\sbr{\E_q \sbr{N \bm{g} \brbth} + \bl \bmu},\\
    \S^{-1}_{t+1} &= \br{1-\beta}\S^{-1}_t +\beta\br{ \E_q\sbr{N H\brbth} +\bl I},
\end{align}
where the expectations can be again evaluated via \acrshort{mc} sampling. By using a single draw $\bth_t \sim \N\br{\bth\vert \bmu_t, \S_t}$, the feasible update reads
\begin{align}
    \bmu_{t+1} &= \bmu_t -\beta \S_{t+1}\sbr{N \bm{g}\br{\bth_t} + \bl \bmu},\\
    \S^{-1}_{t+1} &= \br{1-\beta}\S^{-1}_t +\beta\br{ N H\br{\bth_t} +\bl I}.
\end{align}
To obtain a form for the update that resembles Newton's method where the scaling matrix is estimated online, \citep{khan2018vadam} defines $S_t = \br{\S^{-1}_t -\bl I}/N$ and conversely $\S_t = \br{N\br{S_t +\bl I/N}}^{-1}$, and write the final form of the \acrshort{von} update
\begin{align}
    \bmu_{t+1} &= \bmu_t -\beta \br{S_{t+1}+\frac{\bl}{N} I}^{-1}\br{g\br{\bth_t}+\frac{\bl}{N} \bmu_t},\\
    S_{t+1} &= \br{1-\beta}S_t +\beta H\br{\bth_t}.
\end{align}
Similarly, for a diagonal covariance matrix (thus under a mean-field assumption), with $\sigma^2_t = \sbr{N\br{\bm{s}_t +\frac{\bl}{N}}}^{-1} = \sbr{N\br{\bm{s}_t + \tilde{\bl}}}^{-1}$ and $\bth_t \sim \N\br{\bth\vert \bmu_t,\diag{\sigma^2_t}}$
\begin{align}
    \bmu_{t+1} &= \bmu_t -\beta \br{ \bm{g} \br{\bth_t} +\frac{\bl}{N}\bmu_t} / \br{s_{t+1} + \frac{\bl}{N}} \text{,}\\
    \bm{s}_{t+1} &= \br{1-\beta}\bm{s}_t +\beta\, \diag{H\br{\bth_t}} \text{,}
\end{align}
where the division is intended to be element-wise. Algorithm \ref{alg:VON} summarizes the main elements of \acrshort{von} implementation.

\begin{algorithm}
\caption{\acrlong{von}}
\label{alg:VON}
\begin{algorithmic}
\Repeat
    \State $\sigma = \text{diag}(1/(N(s+\tilde{\bl})^{\frac{1}{2}}))$
    \State Simulate $\bth \sim q_{\bmu,\sigma}$
    \State Randomly sample a data example $\D_i$
    \State $\hat{\bm{g}} = \nabla_\bth \sbr{-\log p\br{\D_i \vert \bth}}$
    \State $\hat{H} = \nabla^2_{\bth\bth}  \sbr{-\log p\br{\D_i \vert \bth}}$
    \State $\bm{s} = \br{1-\beta}\bm{s} +\beta\, \text{diag}\br{H} $
    \State $\bmu = \bmu_t -\beta (\hat{\bm{g}} +\tilde{\bl}\bmu) / (\bm{s} + \tilde{\bl}) $
\Until{stopping condition is met}
\end{algorithmic}
\end{algorithm}

In a mini-batch setting for estimating the stochastic gradient, with $\mathcal{M}$ denoting a mini-batch containing $M$ samples, the stochastic estimates
\begin{align}
  \hat{g}\br{\bth_t} &= \frac{1}{M}\sum_{i \in \mathcal{M}}\nabla_\bth \sbr{-\log p\br{\D_i\vert \bth_t}}\text{,} \\  
  \hat{H}\br{\bth_t} &= \frac{1}{M}\sum_{i \in \mathcal{M}} \nabla^2_{\bth \bth} \sbr{-\log p\br{\D_i\vert \bth_t}}\text{,}
\end{align}
enable the practical implementation of the \acrshort{von} update by replacing $\bm{g}$ and $H$. To make this statement clear, think of $f\brbth$ as the typical negative log-likelihood of a sample (as it is an average across samples), then $\bm{g}$ is the typical gradient for a sample in $\bth$ and, analogously, $H$ is interpreted as the typical (average) value of the Hessian evaluated in $\bth$, resulting when using a single data point. Stochastic gradient estimation estimates $\bm{g}$ by using a single observation $\D_i$ picked at random as an unbiased estimate of the actual gradient of $f\brbth = -\frac{1}{N} \sum_{i=1}^N\log p\br{\D_i\vert \bth}$, $\bm{g} = \frac{1}{N} \sum_{i=1}^n \nabla_\bth \sbr{-\log p\br{\D_i\vert \bth}}$, which would require the parsing of the entire sample. Analogously, one constructs a stochastic estimate of the Hessian with one or $M$ observations (the higher $M$ the lower the variance of the estimator, which is in any case unbiased).

\subsection{\SectionNameWithGlossary{vadam}}
The principle of \acrfull{vadam} is that of augmenting the natural gradient update by incorporating a momentum factor, i.e.,
\begin{equation}
 \bl_{t+1} =\bl_t + \beta \natgrad_\bl \LB +\gamma\br{\bl_t-\bl_{t-1}}
\end{equation}
which slightly extends the form of the update in eq. \eqref{eq:NATGRAD:update_lam}.

Under a Gaussian variational $q$, \citep{khan2018vadam} expresses the momentum update as a \acrshort{von} update with momentum and recovers a variational version of an \acrshort{rmsprop} update, to obtain the following updates
\begin{align}\label{eq:VADAM:heavyball}
    \bmu_{t+1} &= \bmu_t - \bar{\beta}_t \sbr{\frac{1}{\sqrt{\bm{s}_{t+1}}+\tilde{\bl}}}\br{\bm{g}\brbth + \tilde{\bl} \bmu_t} + \frac{\beta_t}{1-\beta_t}\sbr{\frac{\bm{s}_t + \tilde{\bl}}{\bm{s}_{t+1}+\tilde{\bl}}}\br{\bmu_t - \bmu_{t-1}}, \\
    \bm{s}_{t+1} &= \br{1- \bar{\beta}_t} \bm{s}_t + \bar{\beta}_t\br{\bm{g}\brbth}^2\text{,}
\end{align}
where $\bar{\beta}_t = \beta \frac{1-\gamma_1}{1-\gamma_1^t}$, $\bar{\gamma}_t = \gamma_1 \br{1-\gamma_1^{t-1}}\br{1-\gamma_1^t}$ and $\beta$,$\gamma_1$ are learning rates. Note that in the above updates the Hessian is estimated as a squared gradient: details are provided in Section \ref{subsec:VOGN}. These updates can be implemented and used in their actual form, yet they correspond to an \acrshort{adam}-like update. Indeed the above update has the same form of an adaptive version of Polyak's heavy ball method.
\citep{wilson2017marginal} establishes a relation between the form of eq. \eqref{eq:VADAM:heavyball} and the \acrshort{adam} update, and in particular that the \acrshort{adam} update can be written as an adaptive version of the Polyak's heavy ball method.
Upon introducing the typical bias correction terms of \acrshort{adam}, \citep{khan2018vadam} expresses eq. \eqref{eq:VADAM:heavyball} as an \acrshort{adam} update. With respect to a true \acrshort{adam} update, the model weights are stochastically sampled from the posterior, resulting in a Variational version of \acrshort{adam} (\acrshort{vadam}). For the full derivation, which is quite elaborate and extensive, refer to \citep{khan2018fastyetsimple}. 
Algorithm \ref{alg:VADAM} summarizes the \acrshort{vadam} approach for a Gaussian variational posterior with diagonal covariance.

\begin{algorithm}[H]
\caption{\acrlong{vadam}}
\label{alg:VADAM}
\begin{algorithmic}
\Repeat
    \State $\sigma = \text{diag}(1/(N(s+\tilde{\bl})^{\frac{1}{2}})$
    \State Simulate $\bth \sim q_{\bmu,\sigma}\br{\bth}$
    \State Randomly sample a data example $\D_i$
    \State $\hat{g} = \nabla_\bth \sbr{-\log p\br{\D_i \vert \bth}}$
    \State $\bm{m} = \gamma_1 \,\bm{m} +\br{1-\gamma_1}\br{\hat{\bm{g}}+\tilde{\bl}\bmu}$, 
    \State $\bm{s} = \gamma_2 \,\bm{s} + \br{1-\gamma_2}\br{\hat{\bm{g}} \odot \hat{\bm{g}}}$
    \State $\bm{m} = \bm{m}/\br{1-\gamma_1^t}$,
    \State $\bm{s} = \bm{s}/\br{1-\gamma_2^t}$
    \State $\bmu = \bmu -\beta \, \bm{m}/\br{\sqrt{\bm{s}} + \tilde{\bl}}$
\Until{stopping condition is met}
\end{algorithmic}
\end{algorithm}

\subsection{\SectionNameWithGlossary{vogn}}\label{subsec:VOGN}
In the diagonal \acrshort{von} update, the Hessian drives the update for the scaling vector $\bm{s}$ which determines the covariance matrix $\diag{\sigma^2}$. The Hessian can be negative, a situation that could turn $\sigma^2$ negative, which is meaningless. Instead of indirectly tackling the issue by using a constrained optimization approach (which could be difficult to implement), such as a controlled adaptive learning rate, or model reparametrization, \citep{khan2018vadam} proposes the use of the Generalized Gauss-Newton approximation for the Hessian:
\begin{equation}\label{eq:vogn:grad_persample}
    \nabla_{\bth_j\bth_h}^2 f\brbth \approx \frac{1}{M}\sum_{i \in \mathcal{M}} \sbr{\nabla \bth_j f_i\brbth}^2 := \hat{h}_j\brbth.
\end{equation}
This enables a minor but important difference with respect to \acrshort{von}: with an initial positive value for $\sigma^2$, the above approximation will remain positive leading to valid covariance updates. This provides an algorithmic advantage over \acrshort{von} as constraints on $\sigma^2$ are implicitly satisfied. The above implementation of the Hessian estimation, within \acrshort{von}, consists in the \acrfull{vogn} approach \citep{khan2018vadam,osawa2019practical}. The implementation of the above approximation is not immediate as it requires per-sample gradients. The approximation averages squared gradients evaluated on a sample-per-sample basis, as opposed to batch-gradient computation which directly computes the sum of the gradients over mini-batches \citep{osawa2019practical}, which can be seen by comparing \eq{eq:vogn:grad_persample} with \eq{eq:vogn:grad_perminibatch}

The gradient-magnitude approximation that makes use of the mini-batch squared gradient as an approximation for the Hessian,
\begin{equation}\label{eq:vogn:grad_perminibatch}
\nabla_{\bth_j\bth_h}^2 f\brbth \approx \sbr{\frac{1}{M}\sum_{i \in \mathcal{M}} \nabla \bth_j f_i\brbth}^2 = \sbr{\hat{\bm{g}}_j\brbth}^2 \text{,}
\end{equation}
introduces a bias in the Hessian estimation. In fact, increasing the mini-batch size is not advisable as it introduces more bias. Based on the above approximation, \citep{khan2018vadam} advances an \acrshort{rmsprop} version of the \acrshort{von} update.

The practical implementation of \acrshort{vogn} is extensively discussed in \citep{osawa2019practical}, where the efficient implementation of the per-sample gradient computation for certain network layers is discussed: the additional computations needed to access individual gradients bring the run-time within 2-5 times of that of \acrshort{adam}. Algorithm \ref{alg:VOGN} summarizes the implementation of the \acrshort{vogn} optimizer.

\begin{algorithm}[H]
\caption{\acrlong{vogn}}
\label{alg:VOGN}
\begin{algorithmic}
\Repeat
    \State Randomly sample a mini-batch $\mathcal{M}$ of size $M$
    \For{$s=1,\dots,N_s$}
        \State Simulate $\bth_s = \bmu + \eps \sigma$, where $\sigma = (1/(N(\bm{s}+\tilde{\bl})))^\frac{1}{2}$, $\bm{\eps} \sim \N\br{0,I}$
        \For{each $i$ in $\mathcal{M}$}
            \State $\bm{g}_i^{\br{s}} = \nabla_{\bth} \log p\br{\D_i\vert \bth_s}$
        \EndFor
        \State $\hat{\bm{g}}_s = \frac{1}{M}\sum_{i\in \mathcal{M}} \bm{g}_i^{\br{s}}$
        \State $\hat{\bm{h}}_s = \frac{1}{M}\sum_{i\in \mathcal{M}} (\bm{g}_i^{\br{s}})^2$
    \EndFor
    \State $\hat{\bm{g}} = \frac{1}{N_s}\sum_s \hat{\bm{g}}_s$
    \State $\hat{\bm{h}} = \frac{1}{N_s}\sum_s \hat{\bm{h}}_s$
    \State $\bm{m} = \beta_1 \bm{m} +(\hat{\bm{g}} + \tilde{\bl}\bmu)$,
    \State $\bm{s} = (1-\beta_2)\bm{m} + \beta_2 \hat{\bm{h}}$
    \State $\bmu = \bmu -\beta \, \bm{m}(\bm{s} + \tilde{\bl})$
\Until{stopping condition is met}
\end{algorithmic}
\end{algorithm}

The form of Algorithm \ref{alg:VOGN} slightly differs from that of \acrshort{vogn}/\acrshort{adam}. The sampling of the random weights is analogous to that of Algorithm \ref{alg:VADAM} and Algorithm \ref{alg:VON}, yet here posterior samples are built over standard-normal random numbers rather than directly sampling from the multivariate diagonal posterior by the use of the reparametrization trick. Note the index $i$ referring to the individual samples in the mini-batch $\mathcal{M}$. While \acrshort{vogn} uses a single sample for evaluating the stochastic gradients, here $N_s$ draws are averaged to reduce the approximation variance. In particular, the nested for loop computes the single-observation gradient used for the Hessian approximation, each computed in the sampled weight vector $\bth_s$. Draw-specific gradients and Hessian $\hat{\bm{g}}_s$ and $\hat{\bm{h}}_s$ are thus averaged across samples (leading to $\hat{\bm{g}}$ and $\hat{\bm{h}}$) and used in the implementation of the \acrshort{adam}-like update based on momentum (thus the hyperparameters $\beta_1$, $\beta_2$). The pseudo-code in \citep{osawa2019practical} involves an additional tempering parameter and data-augmentation factor along with details for the \acrshort{vogn} parallel implementation, to which we refer for further insights.

\citep{osawa2019practical} furthermore discusses practical implementation aspects typical in \acrshort{ml} such as batch normalization, data augmentation, momentum, and distributed computing. The feasibility of the \acrshort{vogn} update for large-scale experiments with big-data sizes and deep network architectures on standard datasets promotes \acrshort{vogn} as a state-of-the-art method for Bayesian \acrshort{dl}. As a remark, among its limitations, note that \acrshort{vogn} applies to Gaussian variational posteriors with a diagonal covariance matrix only.

\subsection{\SectionNameWithGlossary{qbvi}}\label{subsec:QBVI}
\newcommand{\logpy}{\log p\br{y\vert \bth}}
The \acrshort{bbvi} framework of \citep{ranganath2014black} can benefit from the use of the natural gradients. In fact, in \citep{trusheim2018boosting} natural gradients are estimated via \acrshort{mc} sampling. On the other hand \eq{eq:NATGRAD:update_lam_grad_m} provides an exact framework for computing natural gradients without relying on sampling methods, applicable for the wide class of variational posteriors within the Exponential family, yet model-specific derivations, i.e. the computation of the gradients and Hessian, are involved. The \acrfull{qbvi} approach \citep{magris2022quasi} merges the \acrshort{bbvi} setting with the exact natural gradient computation. \acrshort{qbvi} uses eq. \eqref{eq:NATGRAD:update_lam_grad_m} to turn the computation of the natural gradients into Euclidean gradients of the \acrshort{lb}, which are computed by the use of the score estimation, resembling the \acrshort{bbvi} framework.
On a general level, the \acrshort{qbvi} update estimates the gradient of the \acrshort{lb} with respect to the natural parameters as
\begin{align}
    \natgrad\LB_\bl \br{\bl} &= \natgrad_\bl \Eq{q_\bl}\sbr{\log \frac{p_{\bm{\eta}} \brbth}{q_\bl \brbth}} + \natgrad_\bl\Eq{q_\bl}\sbr{\log p\br{\D\vert \bth}} \label{eq:QBVI:natgrad_general_1}\\
    &=  \bm{\eta} - \bl +\natgrad_\bl\Eq{q_\bl}\sbr{\log p\br{\D\vert \bth}} \text{,} \label{eq:QBVI:natgrad_general_2}
\end{align}
which along with a plain \acrshort{sgd} lead to the update rule
\begin{align}\label{eq:QBVI:update_general}
    \bl_{t+1} = \br{1-\beta}\bl_t + \beta\br{\bm{\eta} + \natgrad_\bl \Eq{q_\bl}\sbr{\log p\br{\D\vert \bth}}} \text{.}
\end{align}
Here the exact computation of the natural gradient is carried out in terms of eq. \eqref{eq:NATGRAD:update_lam_grad_m}, so that the \acrshort{qbvi} update for a generic variational distribution and prior (both within the exponential family) reads, for the natural parameters, as:
\begin{align}\label{eq:QBVI_update}
    \bl_{t+1} &= \br{1-\beta}\bl_t + \beta\br{\eta + \Eq{q_\bl}\sbr{\nabla_{\bm{m}} \sbr{\log q_\bl \brbth}\log p\br{\D\vert \bth}}} \text{.}
\end{align}

Similarly to \citep{khan2018fastyetsimple}, \eq{eq:QBVI:natgrad_general_2} uses the properties of the Exponential family distribution for the prior $p$, with natural parameter $\bm{\eta}$, and $q$ to simplify the first term on the right-side of eq. \eqref{eq:QBVI:natgrad_general_1}. This results in the natural-parameter difference $\bm{\eta}-\bl$, avoiding on the first instance a sampling framework for evaluating the corresponding expectation, i.e. reducing the variance of the estimate for $\natgrad\LB\br{\bl}$, regardless of the estimator used for $\Eq{q_\bl}\sbr{\log p\br{\D\vert \bth}}$.

\citep{magris2022quasi} focuses on the Gaussian variational case, building the \acrshort{qbvi} update on the \acrshort{ngvi} update, but without using the model's gradient and Hessian as for \acrshort{von}. Indeed, using equations \eqref{eq:NGVI:nat_grad_lam1_m1} and \eqref{eq:NGVI:nat_grad_lam2_m2}, \citep{magris2022quasi} recovers, for a full-covariance posterior, the following updates:
\begin{align}
    \iS_{t+1} &= \br{1-\beta}\iS_t +\beta \br{\iS_0  + \Eq{q_\bl}\sbr{\br{\iS_t-\bm{v}_t \bm{v}^\top_t}\log p\br{\D\vert \bth}}} \text{,} \label{eq:QBVI:update_iS} \\
    \bmu_{t+1} &= \bmu_t + \beta\S_{t+1}\br{\iS_0\br{\bmu_0-\bmu_t} + \E{q_\bl}\sbr{\bm{v}_t \log p\br{\D\vert \bth}}} \text{,}\label{eq:QBVI:update_mu}
\end{align}
where $\bm{v}_t = \S^{-1}\br{\bth-\bmu_t}$ and $\bmu_0, \S_0$ denote the mean vector and covariance matrix of the prior distribution on the model parameter $\bth$, respectively. The following naive \acrshort{mc} estimator provides a simple approach for tacking the above expectations
\begin{align}
    \E{q_\bl}&\sbr{\br{\iS_t - \bm{v}_t \bm{v}^\top_t} \logpy }  \\
    &\approx \frac{1}{N_s}\sum_{s=1}^{N_s}\sbr{ (\iS_t - \iS_t\br{\bth_s-\bmu_t}\br{\bth_s-\bmu_t}^\top\iS_t) \log p\br{\D \vert \bth_s}}\text{,}\\
    \E{q_\bl}&\sbr{\bm{v}_t \log p\br{\D\vert \bth} } \approx \frac{1}{N_s} \sum_{s=1}^{N_s}\sbr{\iS_t \br{\bth_s - \bmu_t}\log p\br{\D  \vert \bth_s}} \text{,}
\end{align}
with $\bth_s \sim q_\bl$, $s = 1,\dots,N_s$. Algorithm \ref{alg:QBVI} provides the pseudo-code for the \acrshort{qbvi} implementation.

\begin{algorithm}
\caption{\acrlong{qbvi}}
\label{alg:QBVI}
\begin{algorithmic}
\Repeat
\State Simulate $\bth_s \sim q_{\bl}$, $s = 1,\dots,N_s$
\State $\bm{v} = \iS\br{\bth-\bmu}$
\State $\hat{\bm{g}}_\S = 1/N_s\sum_s \br{\iS_t - \bm{v} \bm{v}^\top} \log p\br{\D \vert \bth_s}$
\State $\hat{\bm{g}}_\bmu =1/N_s\sum_s \bm{v} \log p\br{\D\vert \bth_s}$ 
\State $\iS = \br{1-\beta}\iS + \beta\sbr{\iS_0 + \hat{\bm{g}}_S}$
\State $\bmu = \bmu +\beta \S\sbr{\iS_0\br{\bmu_0-\bmu} + \hat{\bm{g}}_\bmu}$ 
\Until{stopping condition is met}
\end{algorithmic}
\end{algorithm}

\section{Variational Inference on manifolds} \label{subsec:manifold_intro}

In this section, we review a class of methods that pursue a theoretically different approach, i.e., manifold optimization.
The major challenge in \acrshort{vi} optimization is that of guaranteeing constraints on the variational parameter. In a Gaussian, or e.g. an  Inverse Wishart, setting, this corresponds to guaranteeing updates under which the covariance matrix is \acrfull{spd}.

We first introduce in general terms the concept and practice of Riemann optimization.
Therefore, we provide an introduction to Riemann manifolds, the concepts of tangent vectors, tangent spaces, and Riemann gradient to finally provide a more rigorous discussion of the specific problem of performing valid covariance updated for Bayesian Inference under a Gaussian variational \acrshort{ffvi} setting. This section addresses the most crucial aspects concerning the purpose of introducing the \acrlong{mgvb} and \acrlong{emgvb} optimizers. As the topic is itself broad and quite technical, we intentionally provide a descriptive illustration suitable for a general audience, referring to the specialized literature for additional details and a rigorous mathematical treatment at the end of the following section.

\subsection{Introduction to manifold optimization}\label{subsec:Riemann_optimization_intuitive}
Riemann optimization is an alternative to standard \acrshort{sgd} that well fits problems of the kind
\begin{align}
 \argmin_{\bz \in \M} \LB\br{\bz}\text{,}
\end{align}
where $\LB$ is a real-valued function of some parameter $\bz$, defined on a Riemannian manifold $\br{\M,g}$.  A manifold is a topological space that locally resembles Euclidean space near each point, in more detail, is a set that can locally be mapped one-to-one to $\R^k$, where $k$ is the dimension of the manifold. $g$ stands for a metric the manifold is equipped with. 

The optimization problem aims at minimizing $\LB$ by finding the parameter $\bz \in \M$ that lies on the \enquote{smooth surface} of the Riemannian manifold $\br{\M,g}$ resembling a constrained optimization problem requiring the optimum $\bz^*$ to lie on the Riemannian manifold, such as a sphere or the \acrshort{spd} set. As with \acrshort{sgd} in Euclidean vector spaces, Riemann optimization is generally tackled with gradient descent on the surface of the manifold, based on the gradients of $\LB$. Yet, because of the manifold constraint, there are important differences compared to the standard \acrshort{sgd} approach.

The Euclidean vector space $\R^n$ can be interpreted as a Riemannian manifold $\br{\R^n,g}$, with $g$ the common Euclidean metric, where the usual \acrshort{sgd} iteratively updates the parameter $\bz$ as
\begin{equation}
    \bz_{t+1} = \bz_t + \beta\nabla_\bz \LB\br{\bz_t}\text{,} 
\end{equation}
where
\begin{equation}
    \nabla_\bth\LB \br{\bz_t} = \left. \frac{\partial }{\partial \bz}\LB\br{\bz} \right|_{\bz = \bz_t}\text{.}
\end{equation}
It is clear that applying the above to a generic non-Euclidean manifold $\M$ is not trivial as there is no guarantee that $\bz_{t+1}$ is a valid update, i.e. that $\bz_{t+1}$ lies in $\M$. Consider an optimization problem where the parameter $\bz = \br{x,y,z}$ is required to lie on a 2-dimensional spherical manifold of radius 1, embedded in a 3-dimensional ambient space. The Riemannian manifold is $\M=\left\lbrace \bz \in \R^3:\, \vert\vert \bz \vert \vert_2 = 1 \right\rbrace$, with $g$ being the Euclidean metric, and $\LB$ corresponding to a custom loss function for an arbitrary point $\bz$ on the sphere $\M$. Though partial derivatives $\nabla_{\bz}\LB$ are straightforward to compute or evaluate, e.g. with backpropagation, at the current parameter value $\bz_t$, there is no guarantee that the update rule for the Euclidean space $\bz_{t+1} = \bz_t + \beta\nabla_\bz \LB{\bz_t}$ would result in an updated parameter lying on sphere $\M$.
Intuitively, on the \enquote{curved} surfaces of Riemannian manifolds the updates should follow the \enquote{curved} geodesics instead of straight lines as on familiar $\R^n$ Euclidean spaces. To this end, \acrfull{rsgd} constitutes a manifold generalization of the \acrshort{sgd}.

\subsubsection{Elements of Riemannian manifolds}
In $\R^k$, a steepest-ascent approach updates the current iterate $\bz$ in the direction where the first-order increase of the objective function $\LB$ is most positive. Formally, the 
update direction is chosen to be the unit norm vector $\bm{\bm{\eta}}$ that minimizes the directional derivative
\begin{equation}\label{eq:simple_derivative}
 \text{D}\LB\br{\bz} \sbr{\bm{\bm{\eta}}} = \lim_{t\rightarrow 0} \frac{\LB\br{\bz+t \bm{\eta}}-\LB\br{\bz}}{t}\text{.}
\end{equation}
With the domain of $\LB$ being the manifold $\M$, the argument $\bz+t \bm{\bm{\bm{\eta}}}$ does not make much sense in general as $\M$ is not necessarily a vector space.  This leads to the notion of a tangent vector. A possibility for generalizing the directional derivative is to replace $t \mapsto \bz + t\bm{\bm{\eta}}$ by a smooth curve $\gamma$ on $\M$ passing through $\bz$, i.e. $\gamma\br{0} = \bz$. 
A smooth mapping $\gamma:\R \rightarrow \M:\,t\mapsto \gamma\br{t}$ is termed as curve in $\M$. Defining a derivative $\gamma'\br{t}$ as  $\gamma'\br{t} := \lim_{t\rightarrow 0} \frac{\gamma\br{\bz+t}-\gamma\br{\bz}}{t}$ fails on a general manifold as it requires a vector space structure to compute the difference $\gamma\br{\bz+t}-\gamma\br{\bz}$, however for a smooth function  $\LB$ on $\M$ the function $\LB \circ \gamma:\, t \mapsto \LB\br{\gamma\br{t}}$ is a smooth and well-defined function from $\R$ to $\R$ with a well-defined classical derivative. To sum up, let $\bz$ be a point on $\M$, $\gamma$ a curve such that $\gamma\br{0} = \bz$ and  $ \mathcal{F}_\bz\br{\M}$ is the set of smooth real-valued functions defined in a neighborhood of $\bz$ in $\M$. The mapping $\dot\gamma\br{0}$ from $ \mathcal{F}_\bz\br{\M}$ to $\R$ defined by
\begin{equation}  
\dot\gamma\br{0} \LB := \left. \frac{\d}{\d t}\LB\br{\gamma\br{t}} \right|_{t=0} \text{,} \qquad \LB \in \mathcal{F}_\bz\br{\M}
\end{equation}
is called the tangent vector to the curve $\gamma$ at $t=0$.
Note that the above definition defines $\dot\gamma\br{0}$ as a mapping and not as a (e.g. time) derivative as in \eq{eq:simple_derivative}, which would be general meaningless.
We can now formally define the notion of a tangent vector.

A tangent vector $\bxi_\bz$ to a manifold $\M$ at a point $\bz$ is a mapping from $\mathcal{F}_\bz\br{\M}$ to $\R$ such that there exists a curve $\gamma$ on $\M$ with $\gamma\br{0} = \bz$ satisfying 
$$\bxi_\bz \LB := \dot\gamma\br{0} \LB := \left. \frac{\d}{\d t}\LB\br{\gamma\br{t}} \right|_{t=0} \text{,} \qquad \LB \in \mathcal{F}_\bz\br{\M}\text{.}$$
Such a curve $\gamma$ is said to realize the tangent vector $\bxi_\bz$.
The tangent space to $\M$ at $\bz$ is the set of all tangent vectors to $\M$ at $\bz$ and is denoted by $T_\bz \M$. Importantly, it can be shown that $T_\bz \M$ admits a vector space structure, i.e. $T_\bz \M$ is a vector space: it provides a local vector space approximation of the manifold. This property is useful in defining retractions used to locally transform an optimization problem on $\M$ into an optimization problem on the more friendly vector space $T_\bz \M$.
To characterize which direction of motion from $\bz$ produces the steepest increase in $\LB$, to enable a notion of length that applies to tangent vectors, we endow the tangent space $T_\bz \M$ with an inner product $\abr{\cdot,\cdot}$, inducing the norm $\vert\vert \bxi_\bz \vert\vert$ on $T_\bz \M$, from which the direction of the steepest ascent is given by
\begin{equation}
\argmax_{\bxi_\bz \in T_\bz \M:\, \vert\vert \bxi_\bz \vert\vert = 1} \text{D}\LB\br{\bz} \sbr{\bxi_\bz}\text{,}    
\end{equation}
that is, by the unit-norm vector $\bxi_\bz^*$ for which directional derivative $\text{D}$ of $\LB$ in $\bz$ in the direction $\bxi_\bz^*$ is maximized.

A manifold whose tangent spaces are endowed with a smoothly varying inner product is called a Riemannian manifold, and the smoothly varying inner product is called the Riemann metric. With $g$ being such a Riemann metric on $\M$, the Riemannian manifold is, strictly speaking, the couple $\br{\M,g}$. The Euclidean space is the particular Riemannian manifold consisting of a vector space endowed with an inner product.

The gradient of $\LB$ defined on a Riemannian manifold $\M$ at $\bz$ is denoted by the unique element in $T_\bz \M$ that satisfies
\begin{equation}
    \abr{\grad \, \LB\br{\bz},\bxi_\bz} = \text{D}\LB\br{\bz}\sbr{\bxi_\bz}\text{,} \,\,\,\,\, \forall \bxi_\bz \in T_\bz \M \text{.}
\end{equation}
As in correspondence with usual Euclidean gradients, and important in the light of optimization, it can be shown that the direction of  $\grad\, \LB\br{\bz}$ is the steepest ascent direction of $\LB$ at $\bz$
\begin{equation}
\frac{\grad \, \LB\br{\bz}}{\vert\vert \grad \, \LB\br{\bz}\vert\vert} = \argmax_{\bxi_\bz \in T_\bz \M:\, \vert\vert \bxi_\bz \vert\vert = 1} \text{D}\LB\br{\bz} \sbr{\bxi_\bz}  \text{,}
\end{equation}
and that the norm of $\grad \, \LB\br{\bz}$ gives the steepest slope of $\LB$ at $\bz$.

If a manifold $\M_e$ is endowed with a Riemann metric, one would expect that manifolds generated from $\M_e$ inherit its Riemann metric. Let $\M$ be a manifold embedded in $\M_e$ (the subscript $e$ stands for \enquote{embedding}). Since every tangent space $T_\bz \M$ can be regarded as a subspace of $T_\bz \M_e$, the Riemann metric $g_e$ of $\M_e$ induces a Riemann metric $g$ on $\M$ turning $\M $ into a Riemannian manifold. Endowed with this metric, $\M$ is called a Riemannian submanifold of $\M_e$. As it will appear clear in the next section, the submanifold idea is simple yet powerful as any element $\bxi_\bz$ in $T_\bz \M$ can be decomposed into an element of $T_\bz \M$  and its corresponding orthogonal element $\br{T_\bz \M}^\perp$ in $T_\bz \M_e$:
\begin{equation}
\bxi_\bz  = \proj_\bz \bxi_\bz + \proj^\perp_\bz \bxi_\bz \text{,}
\end{equation}
where $\proj_\bz$ denotes the orthogonal projection onto $T_\bz \M$, and $\proj^\perp_\bz$  denotes the orthogonal projection onto $\br{T_\bz \M}^\perp$. 
In this light, by properly defining the embedding ambient space $\M_e$, one may simplify the computation of the Riemannian gradient, and by projection determine the Riemannian gradient in the tangent space $T_\bz \M$ of the manifold $\M$ of interest:
\begin{equation}
\grad\, \LB\br{\bz} = \proj_\bz \grad\,\LB_e\br{\bz},
\end{equation}
with $\LB_e$ being an extended version of the differentiable function $\LB$ defined on $\M_e$ such that its restriction on $\M$ actually coincides with $\LB$.

\begin{figure}
    \centering
    \includegraphics[width=\textwidth]{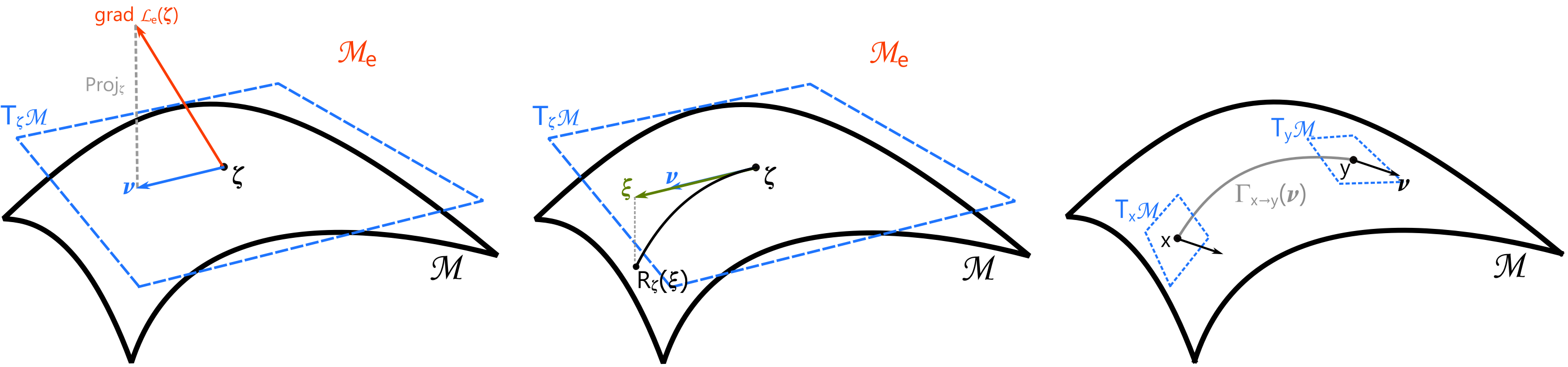}
    \caption{Left: Tangent space and projection of Riemannian gradient. Center: retraction map. Right: vector transport.}
    \label{fig:RSGD}
\end{figure}

Perhaps the most simple tool to tackle Riemann optimization is the \acrfull{rsgd}, first proposed in \citep{bonnabel2013stochastic}. \acrshort{rsgd} typically involves three steps: (i) evaluate the gradient of $\LB_e$ in $T_\bz \M_e$ with respect to $\bz$ at the current value $\bz_t$, Figure \ref{fig:RSGD} (left panel), (ii) project the gradient onto the tangent space of the manifold $\M$ at $\bz_t$, and (iii) update the parameter by performing a gradient step on the surface following the direction of $\grad \, \LB\br{\bz}$, Figure \ref{fig:RSGD} (central panel).

The last step moves the point $\bz_t \in \M$ in the direction of the gradient along a geodesic, onto $\bz_{t+1}$, lying on the manifold. This is achieved by the so-called exponential map, mapping elements from the tangent space to $\M$. The computation of the exponential map is however a cumber-stone in practice: often a first-order approximation is used. Such first-order approximation is called retraction $R_{\bz}\br{\bxi_{\bz}}$, $\bxi_{\bz} \in T_{\bz} \M$. Intuitively, rather than performing an exact update following the curved geodesics of the manifold, retraction first follows a straight line in the tangent space and then orthogonally projects the point in the tangent space on the manifold. Closed-form formulae for retraction on the most common manifold are available in the literature, see e.g. \citep{Absil2009,hu2020brief}, and e.g. \citep{hosseini2015matrix} for the \acrshort{spd} manifold.

The main sources we have used in writing this section are the exhaustive book of \cite{Absil2009}, and the articles \citep{hu2020brief} and \citep{tran_variational_2021}. Classical specialized books on differential geometry are those of \cite{kobayashi1963foundations,do1992riemannian,boothby2003introduction}, while well-suited references for readers without a background in abstract topology are e.g. \citep{tu2011manifolds,do2016differential} and furthermore at an introductory level e.g. \citep{brickell1970differentiable,abraham2012manifolds}. We suggest referring to the literature involved in the above references for further bibliographical details, e.g. the bibliographical notes in Chapter 3 of \citep{Absil2009}. An exhaustive overview of the different applications in manifold optimization in different areas can be found e.g. in \citep{hu2020brief}. For the first developments on \acrshort{sgd} on Riemannian manifolds, we refer to \citep{bonnabel2013stochastic}, further developments towards an RMSprop-like adaptive version of \acrshort{rsgd} can be found in \citep{pmlr-v97-kasai19a}, while Riemann optimization on the lines of the popular Adam and Ada-grad are discussed in \citep{becigneul2018riemannian}. Relevant for the \acrshort{spd} matrix manifold optimization are the results on vector transport and retraction in e.g. \citep{jeuris2012survey} and \citep{sra2015conic}, of remarkable utility for applications. In this regard, we point to \citep{manopt2014} for a manifold optimization package available in multiple languages.

\subsection{Variational Bayes on Riemannian manifolds with natural gradients} \label{subsec:Riemann_optimization_VB}
Variational Bayes on manifolds aims at maximizing the \acrshort{lb} $\LB$ under a fixed-form Gaussian variational posterior guaranteeing a positive-definite form of the covariance matrix $\S$. Thus the variational parameter $\bz$ lies on the Riemannian manifold of \acrfull{spd} matrices $\M = \left\lbrace \S \in \R^{k\times k} : \S = \S^\top, \S \succ0\right\rbrace$. The optimization problem of concern is thus the Riemann optimization problem
\begin{equation}
\argmax_{\bz \in \M} \LB\br{\bz} \text{.}    
\end{equation}
To implement the \acrshort{rsgd} update the manifold $\M$ of \acrshort{spd} matrices is viewed as embedded in the Riemannian manifold $\M_e$. Let $T_\bz \M_e$ be the tangent space to $\M$ at $\bz \in \M_e$. Aligned with the discussion in Section \ref{sec:ExpFamily}, we wish to perform natural gradient updates. To this end, we equip $\M_e$ with the Fisher-Rao metric, defined by the Fisher information matrix $\I_\bz$. 
With such a metric, the inner product between two tangent vectors $\bm{\bm{\nu}}_\bz,\bxi_\bz \in T_\bz\M_e$ is defined as 
\begin{equation}
\abr{\bm{\nu}_\bz,\bxi_\bz} = \bm{\nu}_\bz^\top \I_{\bz}\bxi_\bz \text{,}
\end{equation}
generalizing the usual Euclidean metric $\abr{\bm{\nu}_\bz,\bxi_\bz} = \bm{\nu}_\bz^\top \bxi_\bz$. Let $\LB_e$ be a differentiable function defined on $\M_e$ such that its restriction on $\M$ corresponds to the \acrshort{lb} $\LB$. It can be shown that the steepest ascent direction at $\bz \in \M_e$ for maximizing the objective $\LB_e$ is the natural gradient
\begin{equation}
\natgrad \LB_e\br{\bz} = \iI_\bz \nabla_\bz \LB_e\br{\bz}\text{,}\,\,\,\,\,\, \bz \in \M_e \text{.}
\end{equation}
Note that $\nabla_\bz \LB_e\br{\bz}$ is the usual Euclidean gradient vector of $\LB_e\br{\bz}$, and that, importantly, for $\bz \in \M$,
\begin{equation}
\natgrad_\bz \LB_e\brz = \iI_\bz \nabla_\bz \LB_e\brz = \iI_\bz \nabla_\bz\LB\brz = \natgrad_\bz \LB\brz \text{.}
\end{equation}
That is, the natural gradient of the extended \acrshort{lb} $\LB_e$ in $\M_e$ corresponds to the natural gradient of the \acrshort{lb} on the relevant manifold $\M$.
A framework for formally associating the natural gradient with the Riemannian gradient is provided by the lemma below, see \citep{tran_variational_2021} for more details.

\newtheorem{lemma}{Lemma}
\begin{lemma} The natural gradient of the function $\LB_e$ on the Riemannian manifold $\M_e$ with the Fisher-Rao metric is  the Riemannian gradient of $\LB_e$. In particular, the natural gradient at $\bz$ belongs to the tangent space to $\LB_e$ at $\bz$.
\end{lemma}

This means that with respect to the embedding space $\M_e$, $\natgrad_\bz \LB_e\brz$ is the actual Riemannian gradient, lying on the tangent space $T_{\bz}\M_e$ of $\LB_e$ at $\bz$. Yet we need to associate the Riemannian gradient in $\M_e$ to the \acrshort{lb} $\LB$ in $\M$, the actual objective of \acrshort{rsgd} optimization.

To this end, we naturally equip the submanifold $\M$ with the same Riemann metric inherited from $\M_e$. For $\bm{\nu}_\bz,\bxi_\bz$ now both in $T_\bz\M_e$,
\begin{equation}
\abr{\bm{\nu}_\bz,\bxi_\bz} = \bm{\nu}_\bz^\top \I_{\bz}\bxi_\bz \text{,}
\end{equation}
and we obtain the Riemannian gradient of $\LB$ in $\M$ as the projection of $\grad \LB_e$ on $T_\bz\M$
\begin{equation}
\grad \LB\brz = \text{Proj}_\bz \, \grad \LB_e\br{\bz}\text{.}
\end{equation}

In a Gaussian manifold, $T_\bz\M  \cong T_\bz \M_e$, thus the projection is the identity matrix $I$ and $\grad \, \LB_e = \grad \, \LB$.
Indeed in Gaussian manifolds, $\M$ corresponds to the manifold of \acrshort{spd} matrices whereas $\M_e = \R^{k\times k}$: $T_\bz\M $ and $T_\bz \M_e$ differ by the fact that the first is the tangent space to a certain \acrshort{spd} matrix while the second is the tangent space of a generic $k \times k$ symmetric matrix.
In terms of projection, the difference is irrelevant, thus $\text{Proj}_\bz\grad \, \LB_e = I \grad \, \LB_e$. Mind that, however, on a general level $\text{Proj}_\bz\br{\cdot}$ can be rather difficult to compute. The above relationship between the Riemannian gradient in $\M_e$ and the \acrshort{lb} $\LB$ in $\M$, is established by treating $\M$ as a submanifold of $\M_e$. Alternatively one can derive the Riemannian gradient of $\LB$ requiring $\M$ to be a so-called quotient manifold induced from a Riemannian ambient manifold. In this regard, see \citep{tran_variational_2021} and the references therein.

\acrshort{rsgd} requires a proper retraction $R_\bz:\, T_\bz \M \mapsto \M$ that locally maps $T_\bz\M$ onto the manifold $\M$ while preserving the first-order information of the tangent space in $\bz$. This means that a step of size zero stays at the same point $\bz$ and the differential of the retraction at this origin is the identity mapping \citep{jeuris2012survey}. From the geodesics between two matrices in $\M$, \citep{jeuris2012survey} develops the popular and convenient retraction method (actually a second-order approximation of the exponential map) for the \acrshort{spd} matrices manifold $\M$. This is given by
\begin{equation}\label{eq:retraction}
    R_\bz\br{\bxi} = \bz +\bxi +\frac{1}{2} \bxi \bz^{-1} \bxi \text{,}
\end{equation}
where
\begin{equation}
    \bxi \in T_\bz \M \text{,}
\end{equation}
and updates the current value of $\bz$ on $\M$ by  accounting for $\bxi$ on the tangent space  $T_\bz\M$.

We now add a practically important element to the discussion, vector transport. In order to perform, among the others, the conjugate gradient algorithm, or implement the momentum method within the \acrshort{rsgd} update, we need to relate a tangent vector at some point $\bz \in \M$ to another point $\bm{\bm{\eta}} \in \M$. In differential geometry, this is achieved by a parallel translation, moving tangent vectors from one tangent space to the other, while preserving the length and angle of the original tangent vector, Figure \ref{fig:RSGD} (right panel). 
As for the exponential map, the parallel translation is often approximated by the so-called \textit{vector transport}, which is easier to compute. For $\bxi := \bxi_\bz$ and $\bm{\bm{\eta}} := \bm{\bm{\eta}}_\bz \in T_\bz \M$, an effective vector transform for the manifold of interest is
\begin{align}\label{eq:parallel_transport}
\mathcal{T}_{\bz \rightarrow \bm{\eta}} \br{\bxi} &= Q \bxi Q^\top \text{,} \end{align}
with
\begin{align}
Q &= \bz^{\frac{1}{2}} \exp \br{ \frac{\bz^{-\frac{1}{2}}\bm{\eta} \bz^{-\frac{1}{2}}}{2}} \bz^{-\frac{1}{2}} \text{,}
\end{align}
where $\mathcal{T}_{\bz \rightarrow \bm{\eta}} \br{\bxi} $ denotes the vector transport of the tangent vector $\bxi \in T_\bz \M$ to $\bm{\eta} \in T_{\bm{\eta}}\M$. 
The above vector transport can be written in a compact and computationally advantageous form as (see e.g. \citep{sra2015conic} for details):
\begin{align}
\mathcal{T}_{\bz \rightarrow \bm{\eta}} \br{\bxi} &= E \bxi E^\top \end{align}
with
\begin{align}
E &= \br{\bm{\eta}\bz^{-1}}^{\frac{1}{2}} \,\,\,\,\, \bxi \in T_\bz \M \text{.}
\end{align}
We point out that within the above \acrshort{spd} matrix manifold setting relevant in Gaussian \acrshort{vi}, $\bz$, $\bm{\eta}$, $\bxi$ are matrices and the above equations are well-defined: for homogeneity in notation, we stick with the lower-case bold symbols for indicating elements of a generic space. 

The above vector transport is practically relevant and essential in implementing, e.g., a momentum method on the \acrshort{rsgd} update, that is by using a moving average of the Riemannian gradient at the previous iteration to reduce noise in the estimated gradients and boost convergence:
\begin{equation}
\grad\, \LB\br{\bz_{t+1}}^\text{mom.} := \omega \mathcal{T}_{\bz \rightarrow \bz_{t+1}}\br{\grad\, \LB\br{\bz_{t}}^\text{mom.}} + \br{1-\omega} \grad \, \LB\br{\bz_{t+1}}\text{,}   
\end{equation}
where  $\omega$ is a momentum-weight hyper-parameter.

Manifold optimization in the context of \acrshort{vi} is relatively new, the main reference for this paper is \citep{tran_variational_2021}, whose approach is reviewed in Section \ref{subsec:MGVB}. Besides this, \acrshort{vi} on manifolds is also discussed in \citep{zhou2021manifold,magris2022exact} and appears in \citep{lin2020handling}. Other applications, not related to the purposes of this review, are here not covered, e.g. manifold optimization for variational autoencoders \citep{skopek2019mixed}. Regarding the specific Bayesian inference problem for Neural Networks, at the time of writing, we are not aware of any further works or developments.

\subsection{\SectionNameWithGlossary{mgvb}}\label{subsec:MGVB}

We review the \acrfull{mgvb} method of \citep{tran_variational_2021}. The variational approximation $q_\bl$ to the true posterior is provided by a multivariate Gaussian distribution $\N\br{\bmu,\S}$, $\bmu \in\R^k$. 
The parameter $\bmu,\S$ are jointly collected in the vector $\bz = \br{\bmu,\vec\br{\S}}$, denoting the variational parameter.
There are no restrictions on the structure of the variance-covariance matrix $\S$ which is a generic member of the manifold $\M$ of the \acrfull{spd} matrices, $\M = \left\lbrace \S \in \R^{k\times k} : \S = \S^\top, \S \succ0\right\rbrace$.
 
The exact form of the Fisher information matrix for the multivariate normal distribution is, e.g., provided in \citep{mardia1984maximum} and reads
\begin{equation}
 \I = \begin{pmatrix}
\iS & 0\\
0 & \I\br{\S}
\end{pmatrix} \text{,} 
\end{equation}
with $\I\br{\S}$ being the $k^2 \times k^2$ matrix whose generic element is
\begin{equation}
\I\br{\S}_{\sigma_{ij},\sigma_{kl}} = \frac{1}{2}\text{tr}\br{\iS \frac{\partial \S}{\partial \sigma_{ij}}\iS \frac{\partial \S}{\partial \sigma_{kl}}} \text{.}
\end{equation}
The \acrshort{mgvb} method relies on the approximation $\I\br{\S} \approx \iS \otimes \iS$, where $\otimes$ denotes the Kronecker product. The corresponding approximate inverse \acrshort{fim} reads
\begin{equation}
\iI = \begin{pmatrix}
\S & 0\\
0 & \S \otimes \S
\end{pmatrix}\text{,}
\end{equation}
which leads to a convenient approximate form of the natural gradients of the lower bound with respect to $\bmu$ and $\S$ computed as
\begin{align}
    \natgrad_\bmu \LB &= \S \nabla_\bmu\LB\text{,}\\
    \natgrad_\S \LB & \approx \text{vec}^{-1}\br{\br{\S \otimes \S}\nabla_{\vec\br{\S}}\LB} = \S \nabla_\S \LB \S.
\end{align}
The last equality follows from the fact that for a vector $\bm{v}\in\R^{k\times k}$, $\br{\S\otimes\S}\bm{v} = \vec \br{\S \vec^{-1}\br{\bm{v}} \S}$.
In virtue of the natural gradient definition, the first natural gradient for $\bmu$ is exact while the second one for $\S$ is approximate.
As pointed out in \citep{lin2020handling}, the actual natural gradient for the above Gaussian distribution should read $2\S \nabla_\S \LB \S$, as $\I\br{\S} = 2\iS\otimes\iS$, therefore the \acrshort{mgvb} approximation.
Thus, \citep{tran_variational_2021} adopts the following updates for the parameters of the variational posterior:
\begin{align}
\bmu&= \bmu + \beta \natgrad_\bmu \S\LB  \text{,} \\
\qquad \S &= R_\S\br{\beta\natgrad_\S\LB} \text{,} 
\end{align}
where $R_\S\br{\cdot}$ denotes a suitable retraction for $\S$ on the manifold $\M$, and $\beta$ is the learning rate. Momentum gradients can be used in place of natural ones. In particular \citep{tran_variational_2021} uses retraction in \eq{eq:retraction} and momentum gradients for the updating $\S$. In this regard, \citep{tran_variational_2021} adopts the parallel transport in \eq{eq:parallel_transport} for granting that at each iteration the weighted gradient remains in the tangent space of the manifold $\M$.  

The actual computation of the gradients $\natgrad_\bmu \LB$ and $\natgrad_\S\LB$ boils down to computing $\nabla_\bmu\LB$ and $\nabla_\S\LB$, which in \acrshort{mgvb} is achieved with the black-box estimator
\begin{equation}
    \nabla_\bz\LB\br{\bz} = \Eq{q_\bz}\sbr{\nabla_\bz \sbr{\log q_\bz\brt}\, h_\bz\br\bth}\text{,} 
\end{equation}
where
\begin{equation}
    \quad h_\bz\br{\bth} = \log\sbr{ \frac{p\br{\D\vert \bth}p\br{\bth}}{q_\bz\br{\bth}}}\text{,}
\end{equation}
with $q\sim \N\br{\bmu,\S}$, $\bz = \br{\bmu,\vec\br{\S}}$, and $\LB\br{\bz} \equiv \LB\br{\bmu,\S}$.
In particular, the gradient of $\LB$ with respect to $\bz$ is estimated using $N_s$ samples from the variational posterior through the unbiased estimator 
\begin{equation}\label{eq:naive_mc_grad_estimator}
  \nabla_\bz \LB\br{\bz_t} = \left. \nabla_\bz \LB\br{\bz} \right|_{\bz = \bz_t}\approx  \frac{1}{N_s}\sum_{s=1}^{N_s} \left. \sbr{\nabla_\bz\sbr{ \log q_\bz\br{\bth_s}} \, h_\bz\br{\bth_s}} \right|_{\bz = \bz_t}\text{,}
\end{equation}
where $\bth_s\sim \N\br{\bmu_t,\S_t}$ and the $h$-function is evaluated in the current value of the parameters, i.e. in $\bz_t = \br{\bmu_t,\vec\br{\S_t}}$.
For a Gaussian distribution $q \sim \N\br{\bmu,\S}$ it can be shown that \citep[e.g.][]{wierstra2014natural,magris2022quasi}:
\begin{align}
\nabla_\bmu \log q \br{\bth} &= \iS\br{\bth-\bmu} \label{eq:MC_mu} \text{,}\\
\nabla_\S \log q\br{\bth} &= -\frac{1}{2}\br{\iS - \iS\br{\bth-\bmu}\br{\bth-\bmu}^\top \iS} \label{eq:MC_S} \text{.}
\end{align}
Algorithm \ref{alg:MGVB} summarizes the above process. 
\begin{algorithm}
\caption{\acrlong{mgvb}}
\label{alg:MGVB}
\begin{algorithmic}[1]
\State \text{Simulate: } $\bth_s \sim q_{\bmu,\S}$, $s = 1,\dots,N_s$
\State \text{Compute: } $\hat{\bm{g}}_\bmu$, $\hat{\bm{g}}_{\S}$ 
\State $\bm{m}_\bmu = \hat{\bm{g}}_\bmu$, $\bm{m}_{\iS} = \hat{\bm{g}}_{\S}$ 
\Repeat
\State $\bmu = \bmu + \beta\bm{m}_\bmu $ 
\State $\S_{\text{old}} = \S$,\quad $\S = R_{\S}\br{\beta \bm{m}_{\S}}$
\State \text{Simulate: } $\bth_s \sim q_{\bmu,\S}$, $s =  1,\dots,N_s$ 
\State \text{Compute: } $\hat{\bm{g}}_\bmu$, $\hat{\bm{g}}_{\S}$
\State $\bm{m}_{\bmu} = \omega \bm{m}_\bmu + \br{1-\omega} \hat{\bm{g}}_\bmu$  
\State $\bm{m}_{\S} = \mathcal{T}_{\S_{\text{old}} \rightarrow \S}\br{\bm{m}_{\S}}+\br{1-\omega} \hat{\bm{g}}_{\S}$ 
\Until{stopping condition is met}
\end{algorithmic}
\end{algorithm}

\subsection{\SectionNameWithGlossary{emgvb}}\label{subsec:EMGVB}

The covariance matrix $\S$ is positive definite, its inverse exists and it is as well symmetric and positive definite. Therefore, $\iS$ lies within the manifold $\M$ and can be updated with a suitable retraction algorithm as for $\S$ in Section \ref{subsec:MGVB},
\begin{equation}\label{eq:update_iS_martin}
   \iS = R_{\iS}\br{\beta\natgrad_{\iS}\LB} = R_{\iS}\br{-2\beta\nabla_{\S}\LB}\text{.}
\end{equation}
Opposed to the \acrshort{emgvb} update, relying on the approximation $\iI\br{\S} \approx \iS \otimes \iS$, for tackling a positive-definite update of $\S$, \citep{magris2022exact} targets at updating $\iS$ for which its natural gradient is available in an exact form, by primarily exploiting the duality between the gradients in the natural and expectation parameter space as for \eq{eq:natgrad_lambda_grad_m}, that circumvents the computation and the approximate form of the \acrshort{fim}.

In particular \eq{eq:natgrad_lambda_grad_m} implies that 
\begin{align*}
  \natgrad_{\bmu} \LB  &= \S \nabla_{\bmu} \LB\text{,}  \\
  \natgrad_{\iS} \LB &= -2\natgrad_{\bl_2}\LB = -2\nabla_\S  \text{,}  
\end{align*}
where $\bl_2 =-\frac{1}{2}\iS$ is the second natural parameter of the variational Gaussian posterior $q_\bl$. This leads to the \acrfull{emgvb} updates
\begin{align}\label{eq:updates_EMGVB}
    \bmu_{t+1} &= \bmu_t +  \beta \S \nabla_\bmu \LB_t  \text{,}\\
    \iS_{t+1} &= R_{\iS_t}\br{-2\beta\nabla_{\S}\LB_t}  \text{.}
\end{align}
Despite the approximate \acrshort{mgvb} update for $\S$, \acrshort{emgvb} updates $\iS$ with exact natural gradient computations. Retraction and momentum gradients are computed as in \acrshort{mgvb} but involve $\iS$ in place of $\S$. For retraction,
\begin{equation}
    R_{\iS}\br{\bxi} = \iS +\bxi +\frac{1}{2} \bxi \S \bxi \text{,} \quad \text{where } \quad \bxi \in T_{\iS} \M \text{,}
\end{equation}
with $\bxi$ being the rescaled natural gradient $\beta\natgrad_{\iS}\LB = -2\beta \nabla_\S \LB$. Instead, vector transport reads
\begin{align}
\natgrad_{\iS}^{\text{mom.}}\LB_{t+1} &= \omega\, \mathcal{T}_{\iS_t \rightarrow \iS_{t+1}}\br{\natgrad_{\iS}^{\text{mom.}}\LB_{t}}+\br{1-\omega} \natgrad_{\iS}\LB_{t+1}\text{,}\\
\natgrad_{\bmu}^{\text{mom.}}\LB_{t+1} &= \omega\, \natgrad_{\bmu}^{\text{mom.}}\LB_t + \br{1-\omega}\natgrad_\bmu \LB_t \text{,}   
\end{align}
where the weight $0<\omega<1$ is a hyper-parameter.
As for \eq{eq:QBVI:natgrad_general_2}, by using a Gaussian prior along with a Gaussian posterior, the natural parameter difference becomes particularly simple. With $\bz = \br{\bmu,\S}$,
\begin{align}
    \nabla_\S\Eq{q_\bz} \sbr{\log p\brt - \log q_\bz \brt} &= \frac{1}{2}\iS -\frac{1}{2}\iS_0 \text{,} \label{eq:emgvb:nat_grad_differece_S}\\
    \nabla_\bmu\Eq{q_\bz}\sbr{\log p\brt - \log q_\bz \brt} &= -\iS_0 \br{\bmu-\bmu_0} \label{eq:emgvb:nat_grad_differece_mu}\text{,}
\end{align}
evaluating $\natgrad_\bz \LB$ accounts to practically estimating $\natgrad_\bz \Eq{q_\bz}\sbr{\log p\br{\D\vert \bth}}$ only.
Whether or not one uses the results in \eq{eq:emgvb:nat_grad_differece_S} and \eq{eq:emgvb:nat_grad_differece_mu} under a Gaussian prior assumption, or prefers to use the gradient estimator based on the $h$-function, $h_\bz\brbth =\Eq{q_\bz} \sbr{ \log p\br{\bth}+\log p\br{\D\vert \bth}- \log q_\bz\br{\bth}}$, as in \acrshort{mgvb}, a general-form for the gradients enabling the \acrshort{emgvb} update is provided by
\begin{align}
      \natgrad_{\bmu} \LB \br{\bz_t} &\approx  c_{\bmu_t} + \frac{1}{S}\sum_{s=1}^S \sbr{ \br{\bth_s-\bmu_t} \log f\br{\bth_s} }\label{eq:natgrad_mu_llh}\text{,}\\
      \natgrad_{\iS} \LB\br{\bz_t}  &\approx C_{\S_t} + \frac{1}{S} \sum_{s=1}^S\sbr{\br{\iS_t-\iS_t\br{\bth_s-\bmu_t}\br{\bth_s-\bmu_t}^\top\iS_t} \log f\br{\bth_s} }\text{,} \label{eq:natgrad_S_llh} 
\end{align}
where 
\begin{equation}\label{eq:cases}
  \begin{cases}
    \begin{cases}
        C_{\S_t} = -\iS_t +\iS_0\\
        c_{\bmu_t} = -\S_t \iS_0 \br{\bmu_t-\bmu_0}\\
        \log f\br{\bth_s} = \log p\br{\D\vert \bth_s}
    \end{cases}
    &\text{if prior is Gaussian,}\\\\
    \begin{cases}
        C_{\S_t} = 0\\
        c_{\bmu_t} = \bm{0}\\  
        \log f\br{\bth_s} = h_{\bz_t}\br{\bth_s}
    \end{cases}
    &\text{if prior is Gaussian or not.}
  \end{cases}
\end{equation}
Because of the computations of the constants $C_{\S_t}$ and $c_{\bmu_t}$ under the Gaussian assumption for the prior $p$, the \acrshort{mc} estimators in \eq{eq:natgrad_mu_llh} and \eq{eq:natgrad_S_llh} are of reduced variance. \citep{magris2022exact} also provides analogous simplified updates under the specific assumption that the covariance matrix of $q$ is either diagonal, block-diagonal, or full under an isotropic Gaussian prior whose mean vector is zero and prior covariance matrix $\iS_0$ equal to $\tau I$, with $\tau>0$. Algorithm \ref{alg:EMGVB} summarizes the updating routine.

\begin{algorithm}
\caption{\acrlong{emgvb}}
\label{alg:EMGVB}
\begin{algorithmic}[1]
\State \text{Simulate: } $\bth_s \sim q_{\bmu,\S}$, $s = 1\dots N_s$
\State \text{Compute: } $\hat{\bm{g}}_\bmu$, $\hat{\bm{g}}_{\iS}$ 
\State $\bm{m}_\bmu = \hat{\bm{g}}_\bmu$, $\bm{m}_{\iS} = \hat{\bm{g}}_{\iS}$ 
\Repeat
\State $\bmu = \bmu + \beta m_\bmu $ 
\State $\iS_{\text{old}} = \iS$,\quad $\iS = R_{\iS}\br{\beta \bm{m}_{\iS}}$
\State \text{Simulate: } $\bth_s \sim q_{\bmu,\S}$, $s = 1,\dots,N_s$ 
\State \text{Compute: } $\hat{\bm{g}}_\bmu$, $\hat{\bm{g}}_{\iS}$  
\State $\bm{m}_{\bmu} = \omega \bm{m}_\bmu + \br{1-\omega} \hat{\bm{g}}_\bmu$  
\State $\bm{m}_{\iS} = \mathcal{T}_{\iS_{\text{old}} \rightarrow \iS}\br{\bm{m}_{\iS}}+\br{1-\omega} \hat{\bm{g}}_{\iS}$ 
\Until{stopping condition is met}
\end{algorithmic}
\end{algorithm}

The reader will note that the \acrshort{emgvb} approach is mixing elements of the \acrshort{spd} (matrix) manifold (retraction and parallel transport) with the natural gradient obtained from the Gaussian manifold. A justification for the validity of the above is discussed in \citep{magris2022exact}. The discrepancy between the natural gradient and the Riemannian gradient obtained from the \acrshort{spd} manifold, can be absorbed in the learning rate $\beta$ and the \acrshort{emgvb} update obtained by manifold-consistent derivations from updating $\br{\bmu,2\iS}$. 

\section{Conclusion}

In this survey, we provided an algorithmic overview of standard, as well as, more recently introduced approaches for Bayesian learning for Neural Networks. We structured our description as an easily-accessible introduction to the basic concepts and related methodologies, focused on the core elements and their implementation, providing pseudo-codes and update rules to be used as references for a large number of Bayesian Neural Network implementations.

We provided a foreword introduction to \acrlong{bnn}, their peculiarities, and motivated their use with respect to standard non-Bayesian \acrlong{ann}. In the remainder, we focused on popular and feasible approaches for their estimation. Besides describing some effective \acrlong{mc} methodologies, and introducing \acrlong{mcd} as a Bayesian tool, we presented a variety of methods based on \acrlong{vi} and natural gradients as the main methodological ingredients in modern Bayesian inference for Neural Networks. We presented the widespread \acrlong{bbb} optimizer, followed by two common black-box methods, namely \acrlong{bbvi} and \acrlong{ngbbvi}. Next, we introduced natural gradients and examined the \acrlong{ngvi}, \acrlong{von}, \acrlong{vogn}, and \acrlong{qbvi} approaches. Lastly, by providing an introduction to manifold optimization, we provided a discussion on methods that can implicitly deal with the positive-definite constraint over Gaussian variational specifications, presenting the \acrlong{mgvb} and \acrlong{emgvb} solutions.

We hope that our comprehensive algorithmic treatment of the above-described methodologies will contribute to a better understanding of the connections and differences between the various Bayesian methods for Neural Networks, will support the adoption of such methods in a wide range of applications, and promote further research in this field.

\section*{Acknowledgments}
The research received funding from  the European Union’s Horizon 2020 research and innovation programme under the Marie Sk\l odowska-Curie project BNNmetrics (grant agreement No. 890690).

\bibliographystyle{abbrvnat}
\bibliography{Bibliography}

\begin{thebibliography}{133}
\bibitem[Nielsen(2020)]{nielsen2020note}
F.~Nielsen.
\newblock A note on the natural gradient and its connections with the
  riemannian gradient, the mirror descent, and the ordinary gradient, 2020.
\end{thebibliography}

\clearpage
\appendix
\section{Nomenclature}\label{app:nomenclature}

\begin{table}[h]
  \centering
  \caption{Nomenclature for the most used mathematical symbols.}
  \scalebox{0.8}{
    \begin{tabular}{cccc}
   Symbol & Meaning & Type & Other information \\
    \midrule
      $N$           & Sample size                   & Scalar                   & \\
      $\bx_i$       & $i$-th input sample           & Generally a vector       & \\
      $\by_i$       & $i$-th target                 & Vector or scalar         & Corresponding to $\bx_i$\\
      $\D_x$        & Input Data                    & Generally a matrix       & $\D_x = \cbr{\bx_i}_{i=1}^n$\\
      $\D_y$        & Targets                       & Matrix or vector         & $\D_y = \cbr{\by_i}_{i=1}^n$\\
      $\D$          & Data                          & Matrix                   & $\D = \cbr{\D_y,D_x}$, $\D_i = \cbr{\by_i,\bx_i}$\\
      \multirow{ 3}{*}{$p$}           & \multirow{ 3}{*}{Depends}                       & \multirow{ 3}{*}{p.d.f. or likelihood}     & $p\brbth$ (prior),\\
      &&&$p\br{\bth|\vert \zeta}$ (posterior)\\
      &&&$p\br{\D\vert \theta}$ (likelihood)\\
      $\LB$         & \acrlong{lb}                  & Function                 & \\
      $q$           & Variational posterior         & Vector                   & Generally indexed by the parameter\\
      $\bz, \bl$    & Variational parameter         & Vector                   & $\bz$ generic, $\bl$ natural parameter\\
      $\bm{m}$      & Expectation parameter         & Vector                   & \\
      \multirow{ 2}{*}{$\bth$}        & \multirow{ 2}{*}{Random variable or variable}   & \multirow{ 2}{*}{Vector}                   & Random variable of model's parameter\\
      &&& argument in which e.g. $q$ is evaluated\\
      $\bth_s$      & A sample from $\bth$          & Vector                   & \\
      $k$           & Dimension of the parameter    & Scalar                   & Generally the dimension of $\bth$\\
      $N_s$         & Number of \acrshort{mc} samples  & Scalar                & \\
      $\beta$       & Learning rate                 & Scalar                   & \\
      $\bmu_0,\S_0$ & Prior parameters              & Vector, matrix           & for Gaussian priors \\
      $t$           & Iteration                     & Scalar                   & \\   
      $\I$          & \acrlong{fim}                 & Matrix                   & \\
      $\nabla$      & Euclidean gradient            & Vector, matrix           & \\  
      $\natgrad$    & Natural gradient              & Vector, matrix           & \\  
      $\bxi$        & Element of the tangent space  & Vector, matrix           & \\   
    \bottomrule
    \end{tabular}
    }
  \label{tab:nomenclature}
\end{table}

\clearpage
\printglossaries

\end{document}